\documentclass[conference]{IEEEtran}
\IEEEoverridecommandlockouts
\usepackage{times}
\usepackage[numbers]{natbib}
\usepackage{multicol}
\usepackage[bookmarks=true]{hyperref}
\usepackage{algorithm,algorithmic}
\usepackage{times}

\usepackage{amsmath}
\usepackage{bm}
\usepackage{amssymb}
\usepackage{graphicx}
\usepackage{multirow}
\usepackage{float}
\usepackage{subcaption}

\usepackage{xparse}

\NewDocumentCommand{\subfloat}{ O{} m }{%
  \subcaptionbox{#1}{#2}%
}

\usepackage{booktabs}
\usepackage{siunitx}
\usepackage{makecell}

\usepackage{tabularx}
\usepackage{array}
\newcolumntype{Y}{>{\raggedright\arraybackslash}X}

\usepackage[dvipsnames]{xcolor}
\usepackage{cancel}
\usepackage{stmaryrd}

\usepackage{calrsfs}
\DeclareMathAlphabet{\pazocal}{OMS}{zplm}{m}{n}

\newcommand{\calI}{\pazocal{I}}

\newcommand{\calJ}{\pazocal{J}}

\newcommand{\calK}{\pazocal{K}}

\newcommand{\calL}{\pazocal{L}}

\newcommand{\calM}{\pazocal{M}}

\newcommand{\calP}{\pazocal{P}}

\newcommand{\calU}{\pazocal{U}}

\usepackage{eufrak}

\newcommand{\Rb}{\mathbb{R}}
\newcommand{\Sb}{\mathbb{S}}
\newcommand{\Zb}{\mathbb{Z}}

\newcommand{\vGamma}{{\mathbf{\Gamma}}}

\newcommand{\rT}{{\mathrm{T}}}

\newcommand{\vx}{{\bf x}}

\newcommand{\vI}{{\bf I}}

\newcommand{\vA}{{\bf A}}
\newcommand{\vR}{{\bf R}}

\newcommand{\vL}{{\bf L}}
\newcommand{\vK}{{\bf K}}

\newcommand{\vF}{{\bf F}}
\newcommand{\vM}{{\bf M}}

\newcommand{\vQ}{{\bf Q}}
\newcommand{\vB}{{\bf B}}
\newcommand{\vS}{{\bf S}}
\newcommand{\vX}{{\bf X}}

\newcommand{\vP}{{\bf P}}
\newcommand{\vU}{{\bf U}}

\newcommand{\argmin}{\operatornamewithlimits{argmin}}
\newcommand{\T}{^\mathrm{T}}

\newcommand{\bb}{{\bm b}}

\newcommand{\bd}{{\bm d}}
\newcommand{\be}{{\bm e}}

\newcommand{\bg}{{\bm g}}
\newcommand{\bh}{{\bm h}}

\newcommand{\bk}{{\bm k}}

\newcommand{\bp}{{\bm p}}
\newcommand{\bq}{{\bm q}}
\newcommand{\br}{{\bm r}}
\newcommand{\bs}{{\bm s}}

\newcommand{\bu}{{\bm u}}

\newcommand{\bx}{{\bm x}}
\newcommand{\by}{{\bm y}}
\newcommand{\bz}{{\bm z}}

\newcommand{\bzeta}{{\bm \zeta}}
\newcommand{\blambda}{{\bm \lambda}}

\newcommand{\bnu}{{\bm \nu}}
\newcommand{\bxi}{{\bm \xi}}
\newcommand{\bchi}{{\bm \chi}}

\DeclareFontFamily{OT1}{pzc}{}
\DeclareFontShape{OT1}{pzc}{m}{it}{<-> s * [1.10] pzcmi7t}{}
\DeclareMathAlphabet{\mathpzc}{OT1}{pzc}{m}{it}

\usepackage{amsthm}
\usepackage{soul}

\newtheorem{remark}{Remark}

\newtheorem{problem}{Problem}

\begin{document}

\title{cuNRTO: GPU-Accelerated Nonlinear Robust  Trajectory Optimization}

\author{
\IEEEauthorblockN{
Jiawei Wang\IEEEauthorrefmark{2}\IEEEauthorrefmark{4},
Arshiya Taj Abdul\IEEEauthorrefmark{1}\IEEEauthorrefmark{4},
Evangelos A. Theodorou\IEEEauthorrefmark{1}
}
\IEEEauthorblockA{
\IEEEauthorrefmark{1}Georgia Institute of Technology, Atlanta \quad
\IEEEauthorrefmark{2}University of California, San Diego \quad
\IEEEauthorrefmark{3}Deemos Corporation
}
\IEEEauthorblockA{\IEEEauthorrefmark{4}Equal Contribution}
}


\maketitle

\begin{abstract}
Robust trajectory optimization enables autonomous systems to operate safely under uncertainty by computing control policies that satisfy the constraints for all bounded disturbances. 
However, these problems often lead to large Second Order Conic Programming (SOCP) constraints, which are computationally expensive.
In this work, we propose the CUDA Nonlinear Robust Trajectory Optimization  (cuNRTO) framework by introducing two dynamic optimization architectures that have direct application to robust decision-making and are implemented on CUDA.
The first architecture, NRTO-DR, leverages the Douglas-Rachford (DR) splitting method to solve the SOCP inner subproblems of NRTO, thereby significantly reducing the computational burden through parallel SOCP projections and sparse direct solves. 
The second architecture, NRTO-FullADMM, is a novel variant that further exploits the problem structure to improve scalability using the Alternating Direction Method of Multipliers (ADMM).  
Finally, we provide GPU implementations of the proposed methodologies using custom CUDA kernels for SOC projection steps and cuBLAS GEMM chains for feedback gain updates. 
We validate the performance of cuNRTO through simulated experiments on unicycle, quadcopter, and Franka manipulator models, demonstrating speedups of up to 139.6$\times$. More details are available at \href{https://cunrto.github.io}{cunrto.github.io}.
\end{abstract}

\IEEEpeerreviewmaketitle

\section{Introduction}

Trajectory optimization has become a foundational tool for motion planning and control of robotic systems, enabling robots to compute dynamically feasible paths that minimize objective cost while satisfying constraints \cite{howell2019altro, mastalli2020crocoddyl}. Applications span from manipulation and legged locomotion to aerial vehicles and autonomous driving \cite{bonalli2019gusto, oleynikova2016continuous, saravanos2023distributed}. However, real-world robotic systems inevitably face uncertainty.
When constraints encode safety-critical requirements—such as obstacle avoidance or actuator limits—failing to account for these uncertainties can lead to catastrophic consequences \cite{lew2020chance}.

Handling uncertainty is critical for real-world deployment. One common approach is to model uncertainty as a stochastic disturbance characterized by a probability distribution. 
Existing methods, ranging from chance-constrained optimization \cite{lew2020chance, nakka2023chance} to covariance steering \cite{tsolovikos2021cautious, ratheesh2025operator}, provide probabilistic guarantees on constraint satisfaction. However, stochastic models may not capture all uncertainty types; modeling inaccuracies or exogenous disturbances are often better represented as deterministic uncertainty, referring to disturbances assumed to lie within a bounded set.
In many safety-critical applications, a guarantee is required for \textit{all} possible realizations of such uncertainty. This work addresses trajectory optimization in the presence of these deterministic disturbances.

\begin{figure}[t]
    \centering
    \includegraphics[width=0.475\textwidth]{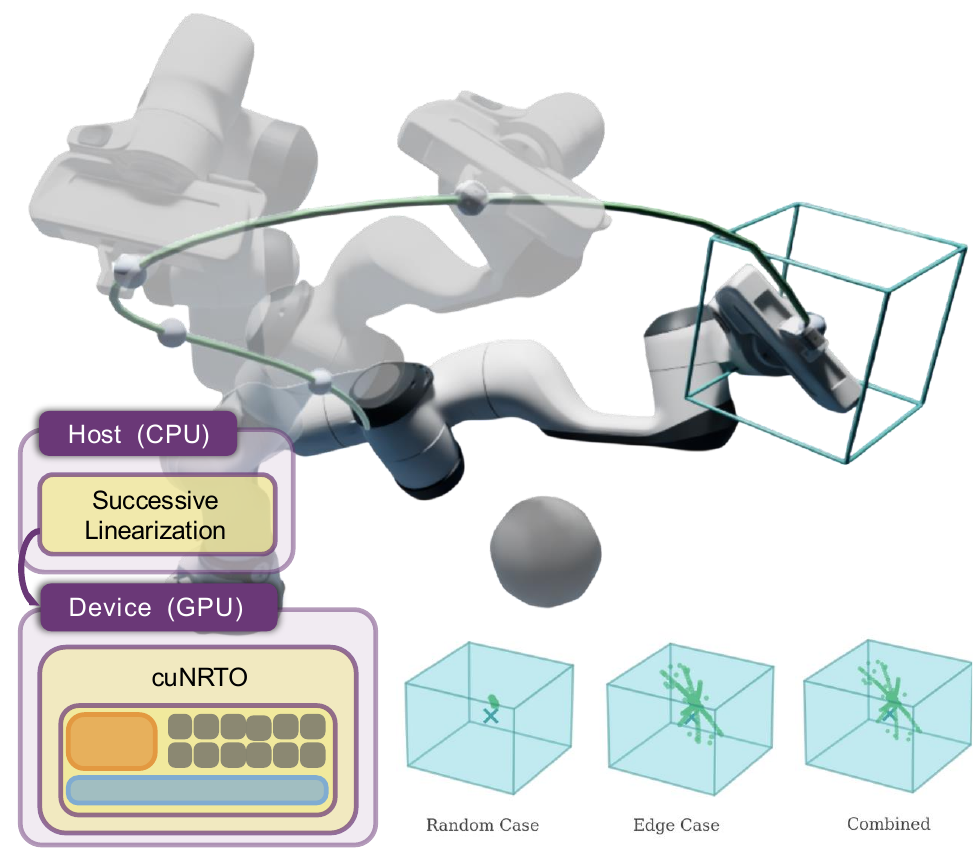}
    \caption{\textbf{cuNRTO on a 7-DoF Franka manipulator:} 
    cuNRTO involves an outer successive linearization (SL) loop run on the host CPU, with an inner loop executed on the GPU.
    Compared to NRTO, cuNRTO achieves a 25.9$\times$ wall-clock speedup on this setting with 100\% constraint satisfaction. The three small boxes show the final state under Monte Carlo rollouts.
    }
    \label{fig:banner}
    \vspace{-0.4cm}
\end{figure} 

Standard approaches to address deterministic uncertainty involve providing worst-case guarantees over bounded uncertainty sets, a concept originating from the field of robust control \cite{safonov2012robust}. This idea has been adapted for trajectory optimization in several forms, ranging from Tube-based and Robust Model Predictive Control (MPC) to min-max methods. Tube-based MPC computes a nominal trajectory and feedback gains to maintain the system within a safe tube; but can be overly conservative due to the decoupling between the nominal control and the feedback gains \cite{mayne2011tube}. Robust MPC typically considers linear systems, and  Min-Max DDP \cite{morimoto2002minimax, sun2018minmax} frame the problem as a game against an adversarial disturbance to optimize a robust policy, failing to address nonlinear state constraints.
While DIRTREL \cite{manchester2017dirtrel} studies robust nonlinear direct transcription under ellipsoidal disturbances with an LQR tracking controller, it assumes uncorrelated uncertainty across the time-steps.
To effectively handle both nonlinear dynamics and state constraints, recent research has shifted toward Robust Optimization (RO) \cite{ben2009robust, bertsimas2011robust}. Some frameworks \cite{DRO_ACC,Kotsalis_covariance_steering} in this category address nonlinear constraints but are limited to linear dynamical systems. 
%
%
Conversely, recent work \cite{aabdulnonlinearcdc} introduced the Nonlinear Robust Trajectory Optimization (NRTO) framework, a trajectory optimization framework that accounts for nonlinearity in both the dynamics and the constraints. 

The NRTO framework comprises three core elements: successive linearization, robust constraint reformulation, and the resolution of linearized subproblem infeasibility via operator-splitting techniques \cite{aabdulnonlinearcdc}. Since the framework enforces robust constraints that must be satisfied for every possible realization of uncertainty within a bounded set, the problem is intractable in its original form due to the infinite number of constraints. To overcome this, the NRTO framework employs robust constraint reformulation, converting the problem into a tractable form that yields Second-Order Conic Programming (SOCP) constraints. 
Despite the effectiveness of this framework, the computational cost of the existing implementation based on Interior Point (IP) methods for solving the SOCP subproblems remains a primary barrier to its widespread adoption for high-dimensional systems with numerous constraints. 

Leveraging GPU acceleration offers a powerful solution to overcome these computational barriers. As single-core CPU performance has largely plateaued, the robotics community has increasingly turned to massively parallel architectures to accelerate complex motion planning tasks. 
Early research in this domain focused primarily on parallelizing rigid body dynamics and their corresponding gradients \cite{plancher2022grid, plancher2021grid_icra}. More recently, GPU acceleration has been pushed into the optimization loop itself.
For example, MPCGPU~\cite{adabag2024mpcgpu} achieves real-time nonlinear MPC by accelerating the dominant sparse Newton solve (PCG) on the GPU. 
Similar GPU-parallelism has also been explored in constrained sampling-based planning ~\cite{wang2025cprrtc}.

Beyond nonlinear MPC, GPU implementations of first-order splitting methods for convex programs, particularly ADMM-based QP solvers such as OSQP~\cite{boyd2011admm,stellato2020osqp}, achieve high throughput by mapping the repeated linear algebra and projection steps to GPU kernels while keeping iterates on-device.
However, the NRTO framework in its current form is not inherently amenable to parallelization. 

This paper presents cuNRTO: a GPU-accelerated nonlinear robust optimization framework by introducing two novel extensions to the original NRTO formulation. This enables efficient parallel execution. Fig.~\ref{fig:banner} provides a high-level overview of the cuNRTO execution flow.  Specifically, the contributions of this paper are three-fold:
\begin{enumerate}
    \item First, we introduce \textbf{NRTO-DR}, a minimal-change methodology to solve the computationally expensive sub-problems of the NRTO framework \cite{aabdulnonlinearcdc} using Douglas-Rachford splitting (DR) \cite{eckstein1992douglas}. The proposed methodology improves the performance over Interior Point (IP) Methods through parallel SOC projections and sparse direct solves that reduce matrix operations.
    %
    \item Second, we propose \textbf{NRTO-FullADMM}, a novel variant of the NRTO framework that could fully leverage GPU acceleration.
     By restructuring the inner ADMM update blocks, this framework integrates SOCP projections directly into the inner loop eliminating extra DR layer and host-device communication overhead.
    %
    %
    \item Third, we develop \textbf{cuNRTO}, which constitutes GPU implementation of the NRTO-DR and NRTO-FullADMM. Our approach utilizes cuBLAS GEMM chains for the feedback gain update and custom CUDA kernels for SOC projections. We validate our approach on unicycle, quadcopter, and Franka manipulator trajectory optimization, demonstrating a speedup of up to 139.60$\times$.
\end{enumerate}
\subsection{Notations}
The space of symmetric positive definite (semi-definite) matrices with dimension $n$ is denoted as $\Sb^{++}_n$ ($\Sb^{+}_n$). The 2-norm of a vector $\bx$ is denoted with $\| \bx \|_2$, while the Frobenius norm of a matrix $\vX$ is given by $\| \vX \|_F$. 
The indicator function of a set $X$, $\pazocal{I}_X$ is defined as $\pazocal{I}(\vx) = \{ 0$ if $\bx \in X$ or $+\infty$ if $\bx \notin X\}$. 
With $\llbracket a, b \rrbracket$, we denote the integer set $\{ [a,b] \cap \Zb \}$.

\subsection{Organization of the paper}
The remainder of this paper is organized as follows. Section II presents the problem statement and a detailed overview of the Nonlinear Robust Trajectory Optimization (NRTO) framework \cite{aabdulnonlinearcdc}. In Sections III and IV, we introduce our core architectural contributions: NRTO-DR and NRTO-FullADMM, respectively. Section V demonstrates the computational performance of these proposed frameworks through a series of simulation experiments involving unicycle, quadcopter, and Franka Emika manipulator models. Finally, Section VI provides concluding remarks and discusses future work.

\section{Nonlinear Robust Trajectory Optimization Framework}
In this section, we outline the nonlinear robust trajectory optimization framework (NRTO) \cite{aabdulnonlinearcdc} considered in this work. We start by presenting the problem statement followed by framework details.
\subsection{Problem Statement}
Consider the following discrete-time nonlinear dynamics
\begin{align}
\bx_{k+1} & = f(\bx_k, \bu_k) + \bd_k, \quad k \in \llbracket 0, T-1 \rrbracket,
\label{original dynamics}
\\
\bx_0 & = \Bar{\bx}_0 + \Bar{\bd}_0,
\label{original dynamics init condition}
\end{align}
where $\bx_k \in \Rb^{n_x}$ is the state, $\bu_k \in \Rb^{n_u}$ is the control input, $f: \Rb^{n_x} \times \Rb^{n_u}  \rightarrow \Rb^{n_x}$ is the known dynamics function and $T$ is the time horizon. The initial state $\bx_0$ consists of a known part $\bar{\bx}_0$, as well as an uncertain part $\bar{\bd}_0$. 
The terms $\bar{\bd}_0, \bd_k \in \Rb^{n_x}$ represent \textit{unknown disturbances}. Considering $\bzeta = [\Bar{\bd}_0; \bd_0; \dots; \bd_{T-1} ] \in \Rb^{(T+1) n_x}$, the uncertainty vector $\bzeta$ is characterized to lie inside a bounded ellipsoidal set defined as follows
\begin{equation}
\begin{aligned}
    \calU[\tau] = 
    \{ \bzeta | ~ \exists  (\bz \in \Rb^{n_z}, & \tau \in \Rb^{+}): 
    \\
    &~
    \bzeta = \vGamma \bz, ~ \bz\T \vS \bz \leq \tau \},
\end{aligned}
\end{equation}
where $\vGamma \in \Rb^{(T+1) n_x \times n_z}$, and $\vS \in \Sb^{++}_{n_z}$. Further, the proposed methodology can be extended for other common types of uncertainty sets such as ellitopes, polytopes, etc. \cite{ben2009robust}.

Consider affine control policies of the following form
\begin{equation}
\begin{aligned}
    & \bu_k = \Bar{\bu}_k + \vK_k \bd_{k-1} 
    \label{control input expression}
\end{aligned}
\end{equation}
where $\Bar{\bu}_k \in \Rb^{n_u}$ is feed-forward control and $\vK_k \in \Rb^{n_u \times n_x}$ are feedback gain. By convention, we set $\bd_{-1} = \bar{\bd}_0$. 

The system is subject to \textit{robust} state and control constraints that should be satisfied for all possible disturbance realizations $\bzeta$ within the uncertainty set $\calU[\tau]$, which are defined as
\begin{align}
\bg(\bx; \bzeta) \leq 0, \quad \forall \bzeta \in \calU[\tau],
\\
\bh(\bu; \bzeta) \leq 0, \quad \forall \bzeta \in \calU[\tau],
\end{align}
where $\bx = [\bx_0; \bx_1; \dots; \bx_T]$, $\bu = [\bu_0; \bu_1; \dots; \bu_{T-1}]$, $\bg: \Rb^{(T+1)n_x} \rightarrow \Rb^{n_g}$, $\bh: \Rb^{Tn_u} \rightarrow \Rb^{n_h}$. 
Further, $g_i$ is concave or linear, and $h_i$ is linear.

Subsequently, the nonlinear robust trajectory optimization problem addressed in this work is presented. 

\begin{problem}[Robust Trajectory Optimization Problem] Find the optimal control policy $\{ \bar{\bu}_k, \vK_k \}_{k=0}^{T-1}$ such that
\begin{equation}
\begin{aligned}
    & \min_{\{\Bar{\bu}_k, \vK_k \}_{k=0}^{T-1}} 
    \sum_{k=0}^{T-1} \Bar{\bu}_k^\rT \vR_u^k \Bar{\bu}_k
    + \| \vR_K^k \vK_k \|_F^2
    \\
    \text{s.t. } \quad
    & 
     \bx_{k+1} = f(\bx_k, \bu_k) + \bd_k,\quad 
     \bx_0 = \Bar{\bx}_0 + \Bar{\bd}_0, 
    \\
    & \bg (\bx; \bzeta) \leq 0, \quad
    \bh (\bu; \bzeta) \leq 0,  
    \quad \forall \bzeta \in \calU[\tau]. \nonumber
\end{aligned}
\end{equation}
\label{problem: original problem}
\end{problem}
%
%
%
%

\subsection{NRTO Algorithm}
In this section, we provide an outline of the NRTO framework \cite{aabdulnonlinearcdc}, which integrates successive linearization, robust constraint reformulation, and Alternating Direction Method of Multipliers (ADMM) \cite{boyd2011admm}. For the simplicity of analysis, we present the problem in terms of the variable $\bk_v \in 
\Rb^{T n_u n_x} $ defined as $
\bk_v = [ \bk_{v,0}; \bk_{v,1}; \dots; \bk_{v,T-1} ]$ with $\bk_{v,k} = \text{vec}(\vK_k)$.

\begin{figure}[t]
    \centering
    \includegraphics[width=0.475\textwidth]{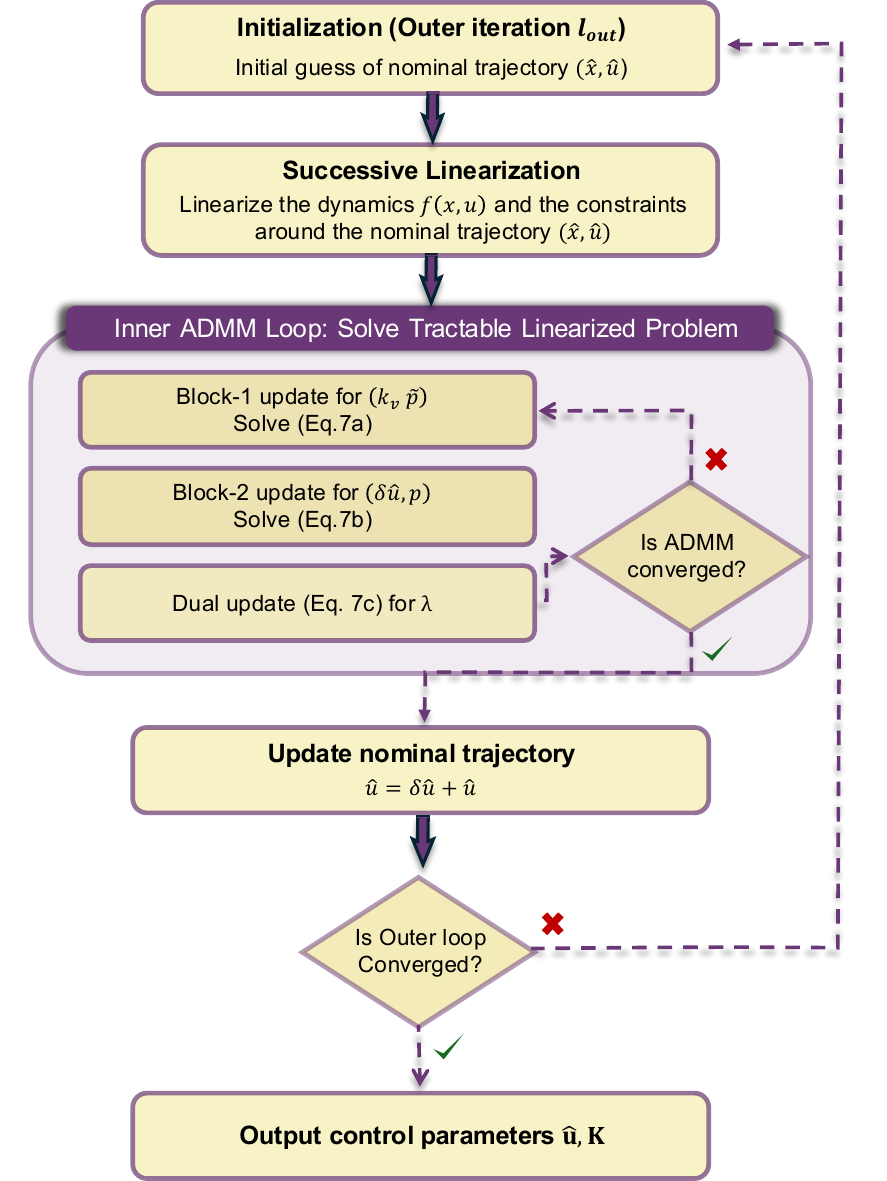}
    \caption{
    \textbf{Overview of NRTO Framework:} 
    A bi-level structure involving an outer successive linearization (SL) loop to generate tractable linearized problem, and an inner ADMM loop to solve the resulting linearized problem.
    }
    \label{fig: NRTO framework}
    \vspace{-0.3cm}
\end{figure}

NRTO is a bi-level algorithm as shown in Fig. \ref{fig: NRTO framework}. In the outer loop of the algorithm, the problem is linearized around a nominal trajectory $\hat{\bu}, \hat{\bx}$ and converted into a tractable form. This results in Second Order Conic Programming (SOCP) constraints as shown below. 

\begin{problem}[Tractable Linearized Problem]
\label{Tractable Linearized Problem}
Find the optimal decision variables $\delta \hat{\bu}, \bk_v, \bp$ such that
\begin{equation*}
\begin{aligned}
    \min_{\delta \hat{\bu}, \bk_v, \bp}\; \;
    & \pazocal{Q}_{\hat{u}} (\delta \hat{\bu})
    + \tilde{\pazocal{Q}} (\bk_v) 
    \\
    \text{s.t.} \quad 
    & \bg^{\text{lin},1}(\delta \hat{\bu}, \bp) \leq 0,
    \\
    & \| \hat{\vA}_j \bk_v + \hat{\bb}_j \|_2 \leq p_j
    \quad j \in \llbracket 1, n_g \rrbracket,
    \\
    & \| \vF_u \delta \hat{\bu} \|_2 \leq r_{\text{trust}}.
\end{aligned}
\label{Linearized Problem - tractable version}
\end{equation*}
where the functions $\pazocal{Q}_{\hat{u}} (\delta \hat{\bu})$, $ \tilde{\pazocal{Q}} (\bk_v)$, $\bg^{\text{lin},1}(\delta \hat{\bu}, \bp)$, 
the matrices $\hat{\vA}_j  \in \Rb^{n_z \times T n_u n_x}$, 
$\hat{\bb}_j \in \Rb^{n_z} $, $\vF_u \in \Rb^{(T+1) n_x \times T n_u }$ are defined in Section I of Supplementary Material (SM). Further, $r_{\text{trust}} \in \Rb^+$ is the trust region radius. 
%

\end{problem}
This linearized problem can be infeasible due to linearization, especially during the initial iterations of the algorithm. Thus, the problem is solved indirectly using ADMM in the inner loop (refer to Algorithm 1 of \cite{aabdulnonlinearcdc} for details). For which, Problem \ref{Tractable Linearized Problem} is converted to the following form solvable using ADMM, by introducing a slack variable $\tilde{p}$,

\begin{problem}[Tractable Linearized Problem - ADMM form]
\label{Tractable Linearized Problem - ADMM}
Find the optimal decision variables $\delta \hat{\bu}, \bk_v, \bp, \tilde{\bp}$ such that
\begin{equation*}
\begin{aligned}
   & \min_{\delta \hat{\bu}, \vK, \bp, \tilde{\bp}} \; \;
    \pazocal{H}_{p} ( \delta \hat{\bu}, \bp )
    + \pazocal{H}_{\tilde{p}} ( \bk_v, \tilde{\bp}  )
    \\
    &~~~ \qquad
    \text{s.t.} \quad \bp = \tilde{\bp}
\end{aligned}
\label{Linearized Problem - admm version}
\end{equation*}
where $\pazocal{H}_{p} ( \delta \hat{\bu}, \bp )$, 
$\pazocal{H}_{\tilde{p}} ( \bk_v, \tilde{\bp}  )$ can be referred from \cite{aabdulnonlinearcdc}.

\end{problem}
The above problem is solved using ADMM by considering the variables $\{ \bk_v, \tilde{\bp} \}$ as the first block, and $\{ \delta \hat{\bu}, \bp \}$ as the second block of ADMM. Each iteration of the ADMM (indexed by $l_{\text{in}}$) involves sequential minimization of the Augmented Lagrangian (AL) of the problem with respect to each block variables, followed by a dual update \cite{boyd2011admm}, given as follows -
\begin{subequations}
\begin{align}
    & \{ \bk_v, \tilde{\bp} \}^{l_{\text{in}}} = \argmin_{\bk_v, \tilde{\bp}} \pazocal{L}_{\rho} (\{ \delta \hat{\bu}, \bp\}^{l_{\text{in}}-1}, \bk_v, \tilde{\bp}; \blambda^{l_{\text{in}}-1} ) 
    \label{NRTO orig - ADMM update step k, ptilde}
    \\
     &
    \{ \delta \hat{\bu}, \bp \}^{l_{\text{in}}}= 
    \argmin_{\delta \hat{\bu}, \bp} \pazocal{L}_{\rho} ( \delta \hat{\bu}, \bp,\{ \bk_v, \tilde{\bp} \}^{l_{\text{in}}};\blambda^{l_{\text{in}}-1} )
    \label{NRTO orig - ADMM update step deltau p}
    \\
   &
    \blambda^{l_{\text{in}}} = \blambda^{l_{\text{in}} - 1} +
    \rho ( \bp^{l_{\text{in}} - 1} - \tilde{\bp}^{l_{\text{in}} - 1} )
    \label{NRTO orig - ADMM update step dual}
\end{align}   
\end{subequations}
where the AL of the problem $\pazocal{L}_{\rho} ( \delta \hat{\bu}, \bp, \bk_v, \tilde{\bp}; \blambda ) = \pazocal{H}_{p} ( \delta \hat{\bu}, \bp )
    + \pazocal{H}_{\tilde{p}} ( \bk_v, \tilde{\bp}  )
    + \blambda\T (\bp - \tilde{\bp})
    + \frac{\rho}{2} \| \bp - \tilde{\bp} \|_2^2$, with $\blambda \in \Rb^{n_g}$ being the dual variable for the constraint $\bp = \tilde{\bp}$ and $\rho >0$ as the penalty parameter.
For the complete details of the NRTO algorithm, the reader is referred to \cite{aabdulnonlinearcdc}.

\subsection{NRTO-LE Algorithm}
This section outlines NRTO-LE \cite{aabdulnonlinearcdc}, an extension of the NRTO framework developed to address linearization error.
While the standard NRTO algorithm ensures that a linearized trajectory satisfies all the constraints, it may not guarantee the safety of the actual trajectory when the linearization error is significant. 
To address this, NRTO-LE models the linearization error at each time step as a bounded uncertainty lying inside an ellipsoidal set. For the comprehensive derivation of the modification, the reader is referred to \cite{aabdulnonlinearcdc}.
\section{ NRTO-DR Framework  }
\label{sec:dr}
In this section, we introduce a framework for accelerating the solving of the sub-problem \eqref{NRTO orig - ADMM update step k, ptilde} which is the most computationally expensive update in the inner loop of the NRTO.
Subsequently, we provide a GPU version of the framework that could further enhance the computational speed.
The design goal of NRTO-DR is to make the smallest algorithmic change to NRTO while replacing the expensive interior-point SOCP subsolve in \eqref{NRTO orig - ADMM update step k, ptilde}. By rewriting this block with Douglas-Rachford splitting, the update decomposes into reusable sparse linear solves and independent SOC projections, exposing the parallelism needed for GPU acceleration.
%
We begin by explicitly writing the ADMM update step \eqref{NRTO orig - ADMM update step k, ptilde}  as follows 
\begin{equation}
\begin{aligned}
    \min_{\bk_v, \tilde{\bp}} \quad &
    \tilde{\pazocal{Q}} (\bk_v)
    + \tfrac{\rho}{2}\big\|\tilde{\bp} - \bp^{l_{\text{in}} -1 } - \blambda^{l_{\text{in}} -1} /\rho \big\|_2^2
    \\
    \text{s.t.}\quad &
    \|\hat{\vA}_j \bk_v + \hat{\bb}_j\|_2 \le \tilde{p}_j,
    \qquad j \in \llbracket 1, n_g \rrbracket
\end{aligned}
\label{eq:dr-subproblem}
\end{equation}

\subsection{Framework}
We introduce a framework leveraging Douglas-Rachford Splitting \cite{lions1979splitting, eckstein1992douglas} (DR) to reduce the computational complexity of solving the above problem. To achieve this, the problem needs to be transformed to a form solvable by DR method. Let us define $\bchi = [\bk_v;\tilde{\bp}]$ and a slack variable $\bs$ such that \eqref{eq:dr-subproblem} can be equivalently given as follows -
\begin{equation}
\begin{aligned}
    \min_{\bchi,\bs}\quad & \tfrac{1}{2}\bchi^\rT \vP \bchi + \bq^\rT \bchi \\
    \text{s.t.}\quad & \vA \bchi + \bs = \mathbf{b}, \qquad \bs \in \calK,
\end{aligned}
\label{eq:dr-standard}
\end{equation}
with $\vP=\mathrm{blkdiag}(\vQ_v,\rho\vI)$ and a product of SOCs $\calK=\calK_1\times\cdots\times\calK_{n_g}$. We provide the explicit construction of $\vA,\mathbf{b}$ and $\calK_i$ in Section II of SM.
The above problem can be rewritten in a form solvable by the DR method as follows -
\begin{equation}
\begin{aligned}
    \min_{\bchi,\bs}\quad & 
    \calJ(\bchi, \bs) + \hat{\calJ} (\bs)
\end{aligned}
\label{Problem: DR version of sub-problem}
\end{equation}
where $\calJ(\bchi, \bs) = (1/2) \bchi^\rT \vP \bchi + \bq^\rT \bchi + \calI_{\vA \bchi + \bs = \mathbf{b}} ( \bchi, \bs)$ and $\hat{\calJ} (\bs) = \calI_{\calK} (\bs)$.

For simplicity, let us also define $\bxi = [ \bchi; \bs ]$.
The update steps in each iteration of relaxed DR method (indexed as $l_{\text{dr}}$) to solve the above problem can be given as follows 
\cite{lions1979splitting, eckstein1992douglas}
\begin{subequations}
\begin{align}
    & \bxi^{ l_{\text{dr}} } = \text{Prox}_{\calJ} ( \tilde{\bxi}^{ l_{\text{dr}} -1 }) 
    \label{DR update step 1}
    \\
    & \bxi_{\text{ref}}^{ l_{\text{dr}} } = 2 \bxi^{ l_{\text{dr}} } - \tilde{\bxi}^{ l_{\text{dr}} -1 }
    \label{DR update step 2}
    \\
    &
    \tilde{\bxi}^{ l_{\text{dr}} } = \tilde{\bxi}^{ l_{\text{dr}} -1} + \alpha( \text{Prox}_{\hat{\calJ}} (\bxi_{\text{ref}}^{ l_{\text{dr}} } ) - \bxi^{ l_{\text{dr}} } )
    \label{DR update step 3}
\end{align}
\end{subequations}
where $\alpha \in (0,1)$ is the relaxation parameter.
We will now simplify the above update steps. The update \eqref{DR update step 1} is a quadratic programming (QP) problem, which can be solved by solving the KKT conditions. This involves solving the following
\begin{subequations}
\begin{align}
    & \begin{bmatrix}
        \bchi^{ l_{\text{dr}} } \\ \by^{ l_{\text{dr}}  }
    \end{bmatrix}
    = 
    (\vL \vL\T)^{-1}
    \begin{bmatrix}
        \vR_\chi \tilde{\bchi}^{ l_{\text{dr}} - 1 } - \bq \\
        \mathbf{b}- \tilde{\bs}^{ l_{\text{dr}} - 1 }
    \end{bmatrix}
    \\
    & \bs^{ l_{\text{dr}}  } = \tilde{\bs}^{ l_{\text{dr}} - 1 } - \vR_s^{-1} \by^{ l_{\text{dr}} }
\end{align}
\label{NRTO-DR 2 update final form}
\end{subequations}
where $\vR_\chi,\vR_s\succ 0$ are fixed proximal scalings, $\vL$ is obtained from the Cholesky decomposition of  $\vK_{\mathrm{KKT}}$. 
The matrix $\vK_{\mathrm{KKT}}$ and the derivation of above update are provided in Section II of the SM. Since $\vK_{\mathrm{KKT}}$ is constant within each iteration of the inner ADMM loop, we factor it once and reuse triangular solves throughout DR. 

\begin{figure}[t]
    \centering
    \includegraphics[width=0.475\textwidth]{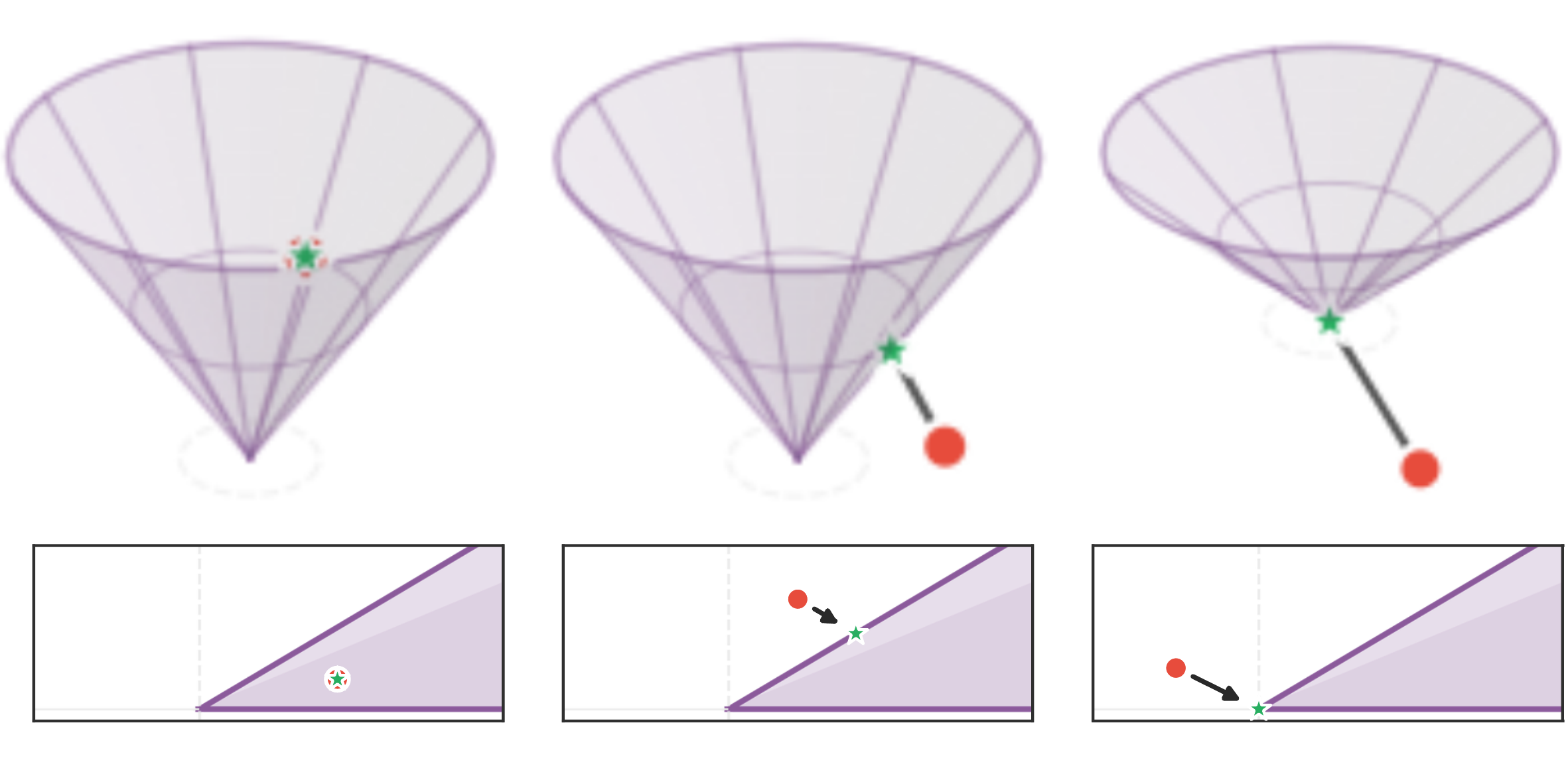}
    \caption{
    %
    %
    Three cases of projecting a point $(\hat{t}, \hat{\by})$ (red) onto a SOC, illustrated in the $(t,\lVert \by \rVert_2)$ plane. 
    \textbf{Left:} if $\lVert \hat{\by} \rVert_2 \le \hat{t}$, the point is already feasible and remains unchanged after projection (green). 
    \textbf{Middle:} if $\lVert \hat{\by} \rVert_2 > |\hat{t}|$, the point lies outside the cone and is projected onto the cone boundary, preserving the direction of $\hat{\by}$. 
    \textbf{Right:} if $\lVert \hat{\by}\rVert_2 \le -\hat{t}$, the point lies in the opposite cone and the projection collapses to the origin.
    }
    \label{fig:SOC projection step}
    \vspace{-0.1cm}
\end{figure}

Now, the update \eqref{DR update step 3} involves
\begin{subequations}
\begin{align}
     & \tilde{\bchi}^{ l_{\text{dr}} } 
     = \tilde{\bchi}^{ l_{\text{dr}} -1 } 
     + \alpha (  \bchi_{\text{ref}}^{ l_{\text{dr}} }  - \bchi^{ l_{\text{dr}} }  )
     \\
     & \tilde{\bs}^{ l_{\text{dr}} }_j 
     = \tilde{\bs}^{ l_{\text{dr}}-1 }_j 
     +  \alpha ( \Pi_{\calK_j} ( \bs_{\text{ref}, j}^{l_{\text{dr}} } ) - \bs^{ l_{\text{dr}} } )
     \quad 
     \forall 
     j \in \llbracket 1, n_g \rrbracket
     \label{NRTO-DR conic projection}
\end{align}
\label{NRTO-DR 3 update final form}
\end{subequations}
where the projection step $\Pi_{\calK_j} ( \bs_{\text{ref}, j}^{l_{\text{dr}} } )$ is defined in Section II of SM, a visual interpretation is provided in Fig. \ref{fig:SOC projection step}.

The DR iterations are terminated when the fixed-point residual $(r_{\text{dr}})$, defined as 
$r_{\text{dr}}^{ l_{\text{dr}} } := \|\tilde{\bs}^{ l_{\text{dr}} }-\tilde{\bs}^{ l_{\text{dr}}  -1}\|_2,$
is below a specified threshold $\varepsilon_{\text{dr}}$ or when the maximum iteration count $L_{\max}$ is reached.

\begin{figure}[t]
    \centering
    \includegraphics[width=0.475\textwidth]{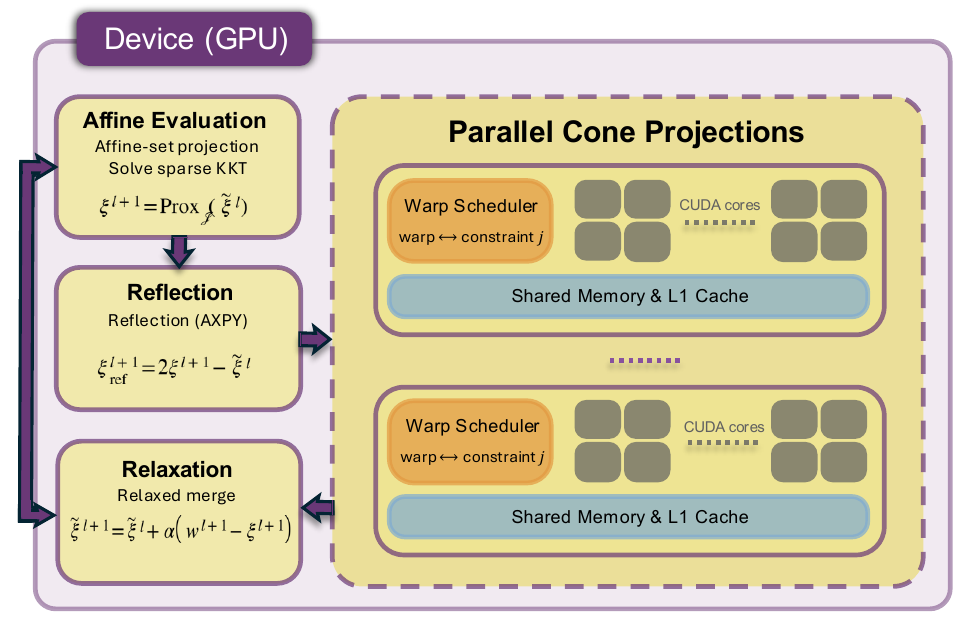}
    \caption{
    \textbf{GPU execution of one relaxed DR iteration:} Each iteration involves an affine-set projection \eqref{DR update step 1}, a reflection step \eqref{DR update step 2}, and massively parallel second-order cone projections \eqref{DR update step 3} that are separable across constraints. 
    The right panel illustrates the GPU mapping: each SOCP constraint block is handled by one warp, the warp scheduler dispatches these warps across streaming multiprocessors, and the grey blocks depict the execution lanes (CUDA cores) that run the warp instructions, with shared memory cache supporting fast projection and vector updates. 
    }
    \label{fig: GPU implementation of DR}
    \vspace{-0.3cm}
\end{figure} 

\subsection{GPU implementation}
The GPU implementation of NRTO-DR (Fig.~\ref{fig: GPU implementation of DR}) follows the relaxed DR updates
\eqref{DR update step 1}-\eqref{DR update step 3}. The affine set proximal step \eqref{DR update step 1} reduces to solving a
fixed sparse KKT system with coefficient matrix $\vK_{\mathrm{KKT}}$, which is factored once across DR iterations.
The details are as follows:
\begin{enumerate}
    \item The one-time sparse factorization of $\vK_{\mathrm{KKT}}$ and per-iteration triangular solves are performed by cuDSS.
    \item SOC projections are computed by a CUDA kernel. The cone-projection case logic is adapted from the open-source SCS implementation~\cite{ocpb:16}.
    \item All vector operations, such as reflection, updates, and residuals, are handled via cuBLAS and kept on-device.
\end{enumerate}


%
%
%
%
%

%
\section{NRTO-FullADMM Framework}
In this section, we present a framework that accelerates the solving of Problem \ref{Tractable Linearized Problem}. Specifically, we propose a novel architecture by modifying the inner ADMM loop of the NRTO framework.
We start by rewriting Problem \ref{Tractable Linearized Problem} by introducing a slack variable $\bnu$, as follows 
\begin{subequations}
\begin{align}
    \min_{\delta \hat{\bu}, \bk_v, \bnu,  \bp, \tilde{\bp}}\; \;
    & \pazocal{Q}_{\hat{u}} (\delta \hat{\bu})
    + \tilde{\pazocal{Q}} (\bk_v) \nonumber
    \\
    \text{s.t.} \quad 
    & \bg^{\text{lin},1}(\delta \hat{\bu}, \bp) \leq 0,
    ~~
    \| \vF_u \delta \hat{\bu} \|_2 \leq r_{\text{trust}}
    \label{Approach_2 uhat constraints}
    \\
    &   
    \| \bnu_j \|_2 \leq \tilde{p}_j 
    ~~
    j \in \llbracket 1, n_g \rrbracket,
    \label{Approach_2 nuK constraints}
    \\
    & \bp = \tilde{\bp}, 
    ~~
    \hat{\vA}_j \bk_v + \hat{\bb}_j = \bnu_j, ~
    j \in \llbracket 1, n_g \rrbracket
    \label{Approach_2 coupling constraints}
\end{align}
\label{NRTO-FullADMM problem form}
\end{subequations}
\vspace{-2em}
\subsection{Framework}
We propose a framework to solve the above problem using a scaled form of ADMM \cite{boyd2011admm}. The variables $(\bnu, \tilde{\bp})$ are considered as the first block, and $(\delta \hat{\bu}, \bp, \bk_v)$ are as the second block of the inner ADMM loop, with \eqref{Approach_2 coupling constraints} being the coupling constraints. 
%
The AL of the above problem is disclosed in Section III of SM, with the penalty parameter as $\rho$, and $\rho \blambda_{p}, \rho \blambda_{\nu}$ as the dual variables.
Subsequently, we present the update steps in the $(l_{\text{in}}^{th})$ iteration, with the detailed derivation disclosed in Section III of SM.
\paragraph{Block-1 update}
This update can be decoupled with respect to the variables $\{ \bnu_j, \tilde{p}_j \}_{j = 1}^{n_g}$, and is given as follows 
\begin{equation}
\begin{aligned}
     & (\bnu^{l_{\text{in}} }_j, \tilde{p}_j^{ l_{\text{in}} })  
     \\
    & ~~~~~ =
    \Pi_{\text{SOC}} ( \hat{\vA}_j \bk_v^{l_{\text{in}} -1 } + \hat{\bb}_j + \blambda_{\nu, j}^{l_{\text{in}} -1 },  p_j^{l_{\text{in}} -1 } + \lambda_{p,j}^{l_{\text{in}} -1 } )
\end{aligned}
\label{NRTO-FullADMM B-1 update}
\end{equation}
where $\Pi_{\text{SOC}}(\hat{\by},\hat{t})$ represents projection of $(\hat{\by},\hat{t})$ onto the cone $\| \by \|_2 \leq t $ (shown in Fig. \ref{fig:SOC projection step}) . This projection step is similar to the projection involved in \eqref{NRTO-DR conic projection} of NRTO-DR framework.
\paragraph{Block-2 update}
The variables $(\delta \hat{\bu}, \bp)$ and $\bk_v$ are decoupled, with the ADMM update steps given as 
\begin{subequations}
\begin{align}
    (\delta \hat{\bu}^{l_{\text{in}}}, \bp^{ l_{\text{in} }})
    & = 
    \argmin_{\delta \hat{\bu}, \bp } \pazocal{Q}_{\hat{u}} (\delta \hat{\bu})
    + \frac{\rho}{2} \| \bp - \tilde{\bp}^{l_{\text{in}}} + \blambda_p^{l_{\text{in}} - 1} \|_2^2 \nonumber \\
    & ~~~~~~~
    \text{s.t. }  \eqref{Approach_2 uhat constraints} 
    \label{Approach_2 ADMM B2 update uhat}
    \\
    \bk_v^{l_{\text{in}}} 
    & = \bq + \calM \bk_v^{l_{\text{in}} - 1}  
    + \bar{\calM} \bnu^{l_{\text{in}}}
    \label{Approach_2 ADMM B2 update kv}
\end{align}
\label{NRTO-FullADMM block-2 update}
\end{subequations}
%
where $\bq, \calM, \bar{\calM} $, disclosed in Section III of SM, remain constant throughout the inner ADMM loop. 
The update \eqref{Approach_2 ADMM B2 update uhat} is similar to the update step \eqref{NRTO orig - ADMM update step deltau p} of NRTO. 
\paragraph{Dual update}
The update of the regularized dual variables is given as 
\begin{subequations}
 \begin{align}
    \blambda_p^{l_{\text{in}}} & =
    \blambda_p^{l_{\text{in}} -1 }
    + (\bp^{l_{\text{in}}} - \tilde{\bp}^{l_{\text{in}}}) 
    \\
    \blambda_{\nu, j}^{l_{\text{in}}} & =
    \blambda_{\nu, j}^{l_{\text{in}} -1 }
    + ( \hat{\vA}_j \bk_v^{l_{\text{in}}} + \hat{\bb}_j - \bnu^{l_{\text{in}} }_j ) 
\end{align}   
\label{NRTO-FullADMM dual updates}
\end{subequations}
%
%
The inner ADMM loop is terminated when the primal residual $r_p := \| \mathbf{p}^{l_{\text{in}}} - \tilde{\mathbf{p}}^{l_{\text{in}}} \|_2$ and the dual residual 
$r_d := \rho \| \tilde{\mathbf{p}}^{ l_{\text{in}} } - \tilde{\mathbf{p}}^{ l_{\text{in}} -1 }\|_2$ fall below their respective thresholds, $\varepsilon_p$ and $\varepsilon_d$, or 
when max iterations $L_{\max}$ are reached.


 %
\begin{algorithm}[t]
\caption{ Inner ADMM Loop  - NRTO-FullADMM}
\label{alg:FullADMM}
\begin{algorithmic}[1]
\REQUIRE $\{ \hat{\vA}_j, \hat{\bb}_j \}_{j =1}^{n_g}$, tol $\varepsilon$, max iters $L_{\max}$
\STATE Form $\bq, \calM, \{ \bar{\calM}_j \}_{j = 1}^{n_g}$
\STATE Initialize $\blambda_p^0, \blambda_{\nu}^0, \bk_v^0, \bp^0
\ \gets \mathbf{0}$
\FOR{$ l_{\text{in}} =0 $ to $L_{max}$}
\STATE 
Sequentially run \eqref{NRTO-FullADMM B-1 update}, \eqref{NRTO-FullADMM block-2 update}, \eqref{NRTO-FullADMM dual updates} 
\STATE \textbf{if} ($r_p \leq \varepsilon_p$ and $r_d \leq \varepsilon_d $) \textbf{then break}
\ENDFOR
\STATE \RETURN $( \delta \hat{\bu}, \bk_v, \bnu,  \bp, \tilde{\bp} )$ 
\end{algorithmic}
\label{NRTO-FullADMM}
\end{algorithm}

\begin{figure*}[!h]
    \centering
    \includegraphics[width=\textwidth]{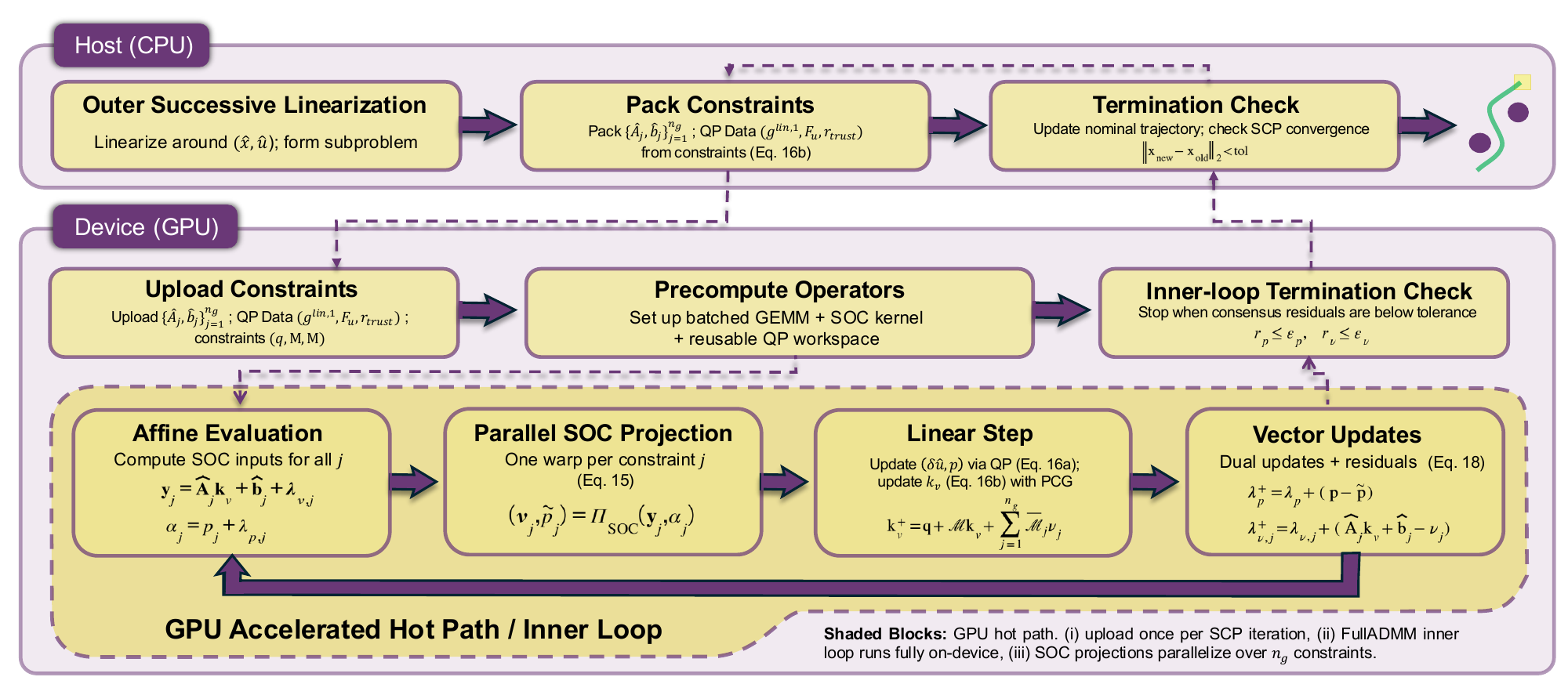}
    \caption{
    \textbf{cuNRTO pipeline for NRTO-FullADMM:} Each outer SL iteration on the host CPU linearizes the problem, packs the SOCP and QP data, and uploads constants to the GPU once. The FullADMM inner loop runs entirely on-device: (i) batched affine evaluation forms per-constraint inputs, (ii) SOC projections are computed in parallel over $j$, (iii) Block-2 updates solve the QP and update $\bk_v$ using prepacked operators, and (iv) dual updates and residual checks determine termination.}
    \label{fig:FullADMM GPU}
\end{figure*}

\subsection{GPU Implementation}
An overview of the end-to-end on-device inner ADMM loop pipeline for NRTO-FullADMM is shown in Fig.~\ref{fig:FullADMM GPU}. 

The details are as follows
%
\subsubsection{On-device data layout}
We store $\bnu \in \Rb^{n_g\times n_z}$ and $\blambda_\nu \in \Rb^{n_g\times n_z}$ as contiguous row-major blocks with each row corresponding to one constraint $j$, and $\bp,\tilde{\bp},\blambda_p \in \Rb^{n_g}$ as contiguous vectors.
All constant matrices $\{ \hat{\vA}_j, \hat{\bb}_j\}_{j=1}^{n_g}$ as well as $\bq,\calM, \bar{\calM}$ are uploaded once per outer iteration.

\subsubsection{Block-1}
The Block-1 update requires projecting
\[
(\hat{\by}_j,\hat{\alpha}_j) \;=\; 
(\hat{\vA}_j\bk_v + \hat{\bb}_j + \blambda_{\nu,j},\; p_j + \blambda_{p,j})
~~ \text{onto} ~~
\|\bnu_j\|_2 \le \tilde p_j.
\]

We implement this in two GPU steps. First, affine evaluation. We compute all $\hat{\by}_j$ via a sequence of cuBLAS GEMM calls that exploit the structured factorization of $\hat{\vA}_j$ through $\vM$, $\vU_k$, and $\vB_k$, avoiding explicit dense matrix formation. Second, SOC projection. We launch a CUDA kernel that performs the SOC projection for each constraint block in parallel. The projection routine follows the standard SOC projection cases and is adapted from the SCS reference implementation~\cite{ocpb:16}, with modifications for warp-level execution and our batched memory layout.
This step is bandwidth-bound and scales linearly with $n_g$, it achieves near-ideal parallel efficiency.

\subsubsection{Block-2}
The update \eqref{Approach_2 ADMM B2 update uhat} is a convex QP with fixed quadratic term and fixed constraints \eqref{Approach_2 uhat constraints}. 
We solve it using preconditioned conjugate gradient (PCG) with a Jacobi preconditioner. Crucially, its linear algebra is reusable across inner iterations, so any preconditioner is computed once per outer iteration and reused thereafter. The $\bk_v$ update \eqref{Approach_2 ADMM B2 update kv} is solved via PCG with matrix-free Hessian-vector products, avoiding explicit formation of the dense Hessian matrix.

\subsubsection{Dual updates}
Dual updates are vector axpy-style operations. We compute the primal residual $r_p$ and dual residual $r_d$ on the GPU and terminate when both fall below the tolerance.

\begin{remark}
    NRTO-FullADMM scales effectively over NRTO and NRTO-DR through two key innovations. First, it replaces the nested structure of NRTO-DR by directly integrating the parallel SOCP block projections into the inner ADMM updates. Second, it eliminates CPU-GPU data transfer bottlenecks by executing the entire inner ADMM loop on the GPU.
\end{remark}

\begin{figure*}[!ht]
    \centering
    \begin{subfigure}[t]{0.32\textwidth}
        \centering
        \includegraphics[width=\linewidth]{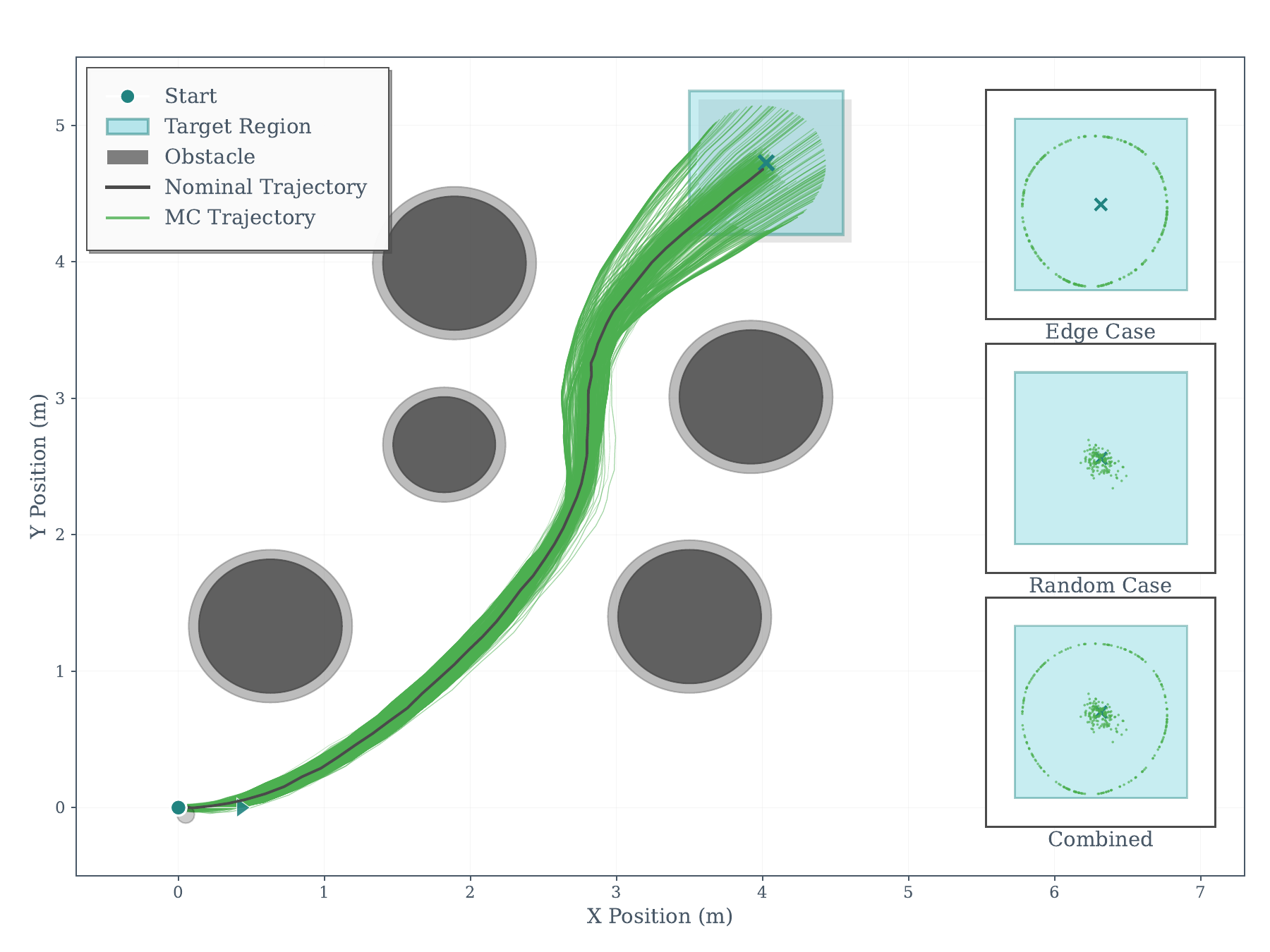}
        \caption{NRTO}
        \label{fig:demo1}
    \end{subfigure}\hfill
    \begin{subfigure}[t]{0.32\textwidth}
        \centering
        \includegraphics[width=\linewidth]{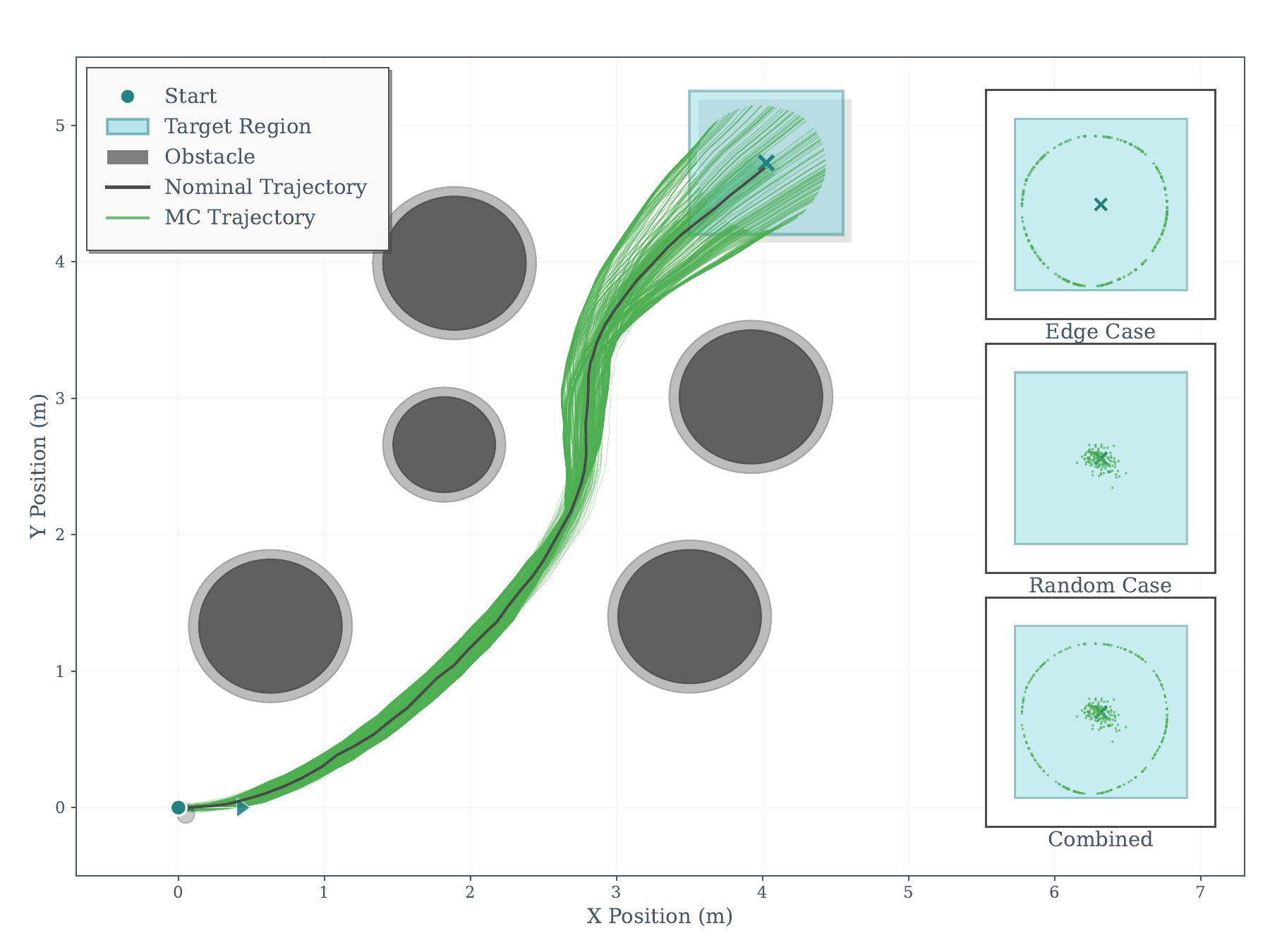}
        \caption{NRTO-DR}
        \label{fig:demo2}
    \end{subfigure}\hfill
    \begin{subfigure}[t]{0.32\textwidth}
        \centering
        \includegraphics[width=\linewidth]{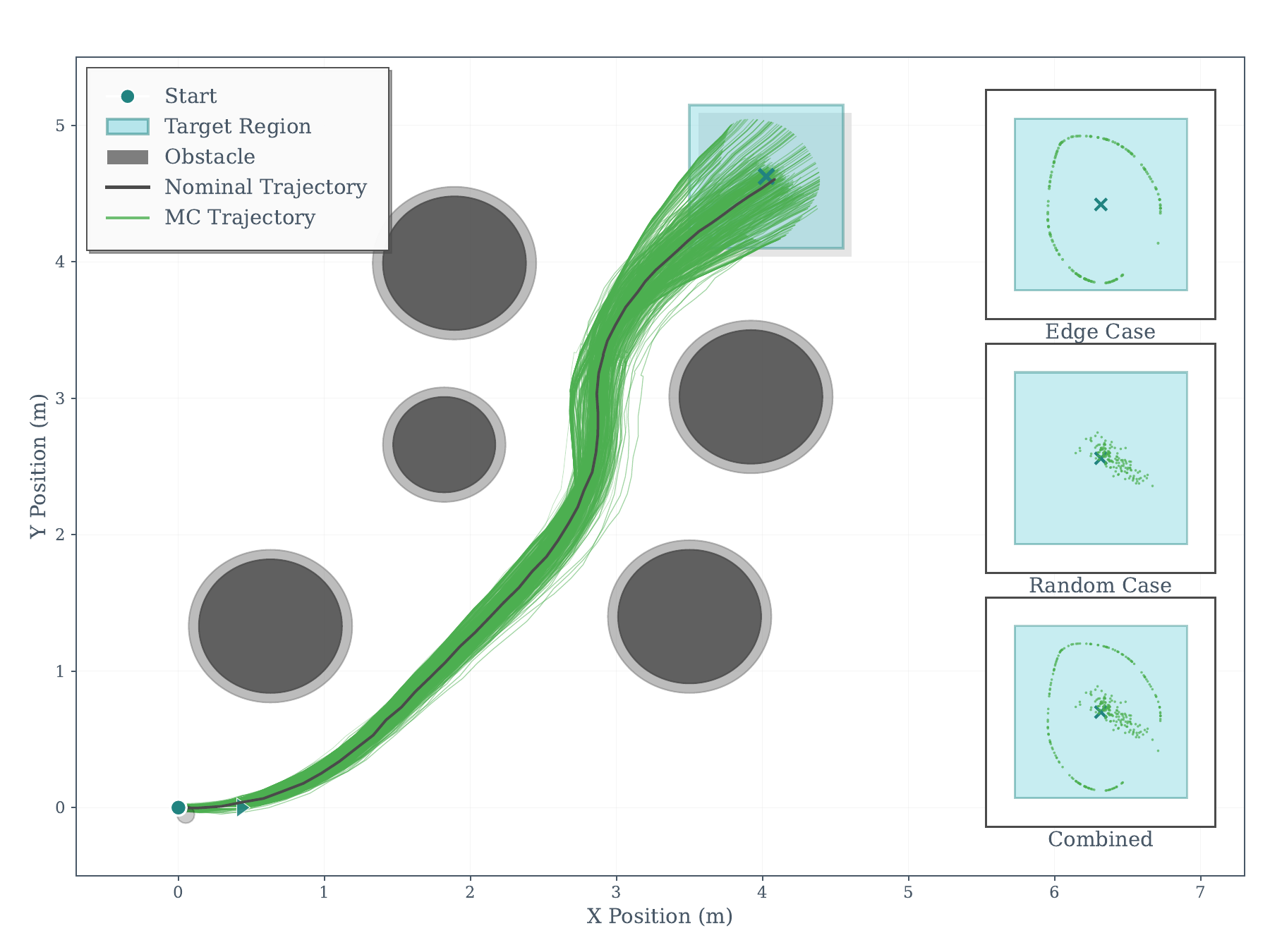}
        \caption{NRTO-FullADMM}
        \label{fig:demo3}
    \end{subfigure}

    \caption{
    \textbf{Performance comparison on Unicycle model with five obstacles:}
    We compare (a) the baseline NRTO solver, (b) NRTO-DR, and (c) NRTO-FullADMM. All produce collision-free trajectories that satisfy the robust constraints for 2,000 disturbance realizations. The right insets represent the distribution of terminal state realizations for random, edge, and combined rollouts, demonstrating robustness. 
    }
    \label{fig:exp_unicycle}
\end{figure*}

\begin{figure*}[!ht]
    \centering
    \begin{subfigure}[t]{0.48\textwidth}
        \centering
        \includegraphics[width=\linewidth]{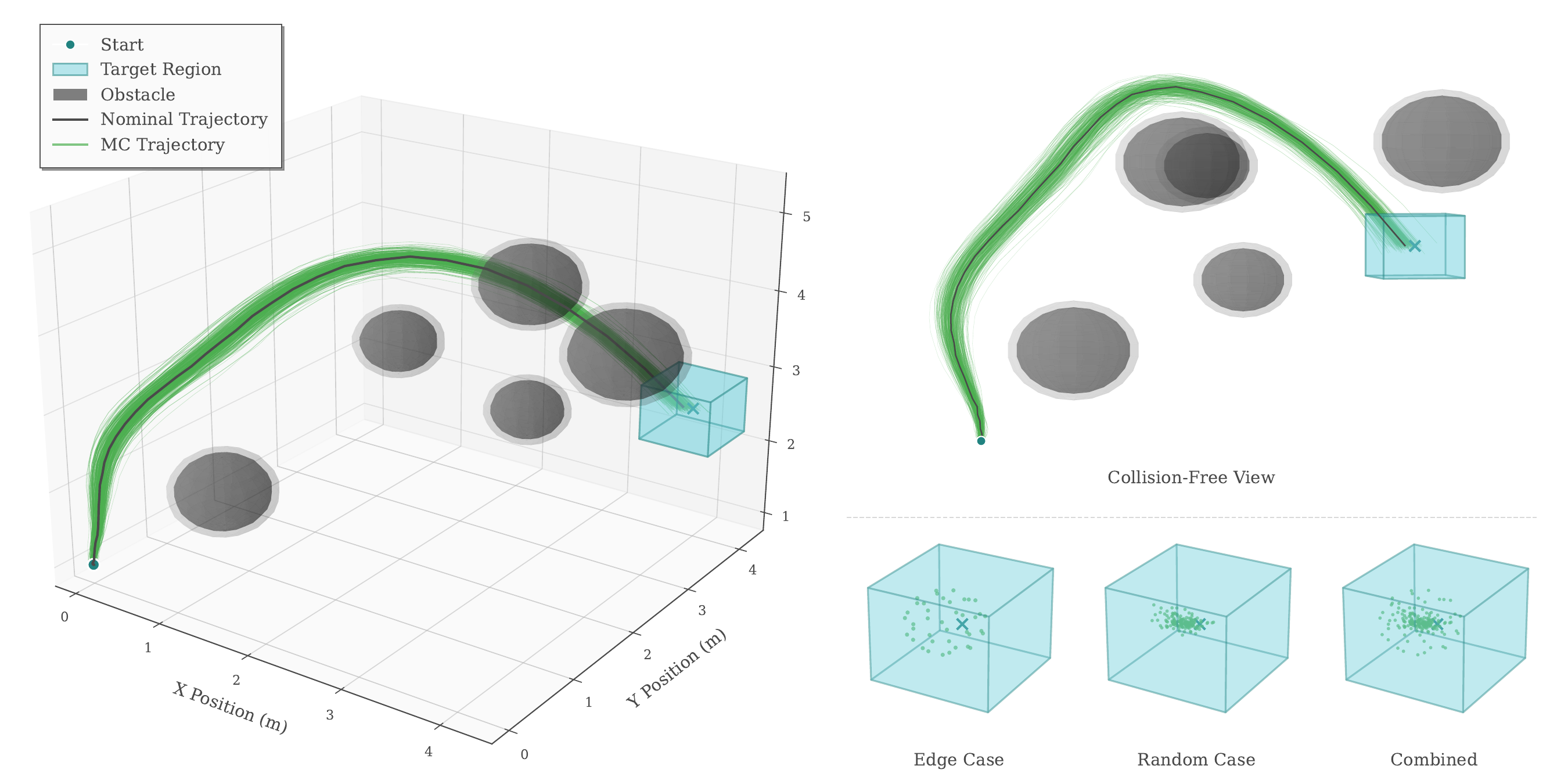}
        \caption{NRTO-DR}
        \label{fig:quadcopter_nrto_dr_demo}
    \end{subfigure}\hfill
    \begin{subfigure}[t]{0.48\textwidth}
        \centering
        \includegraphics[width=\linewidth]{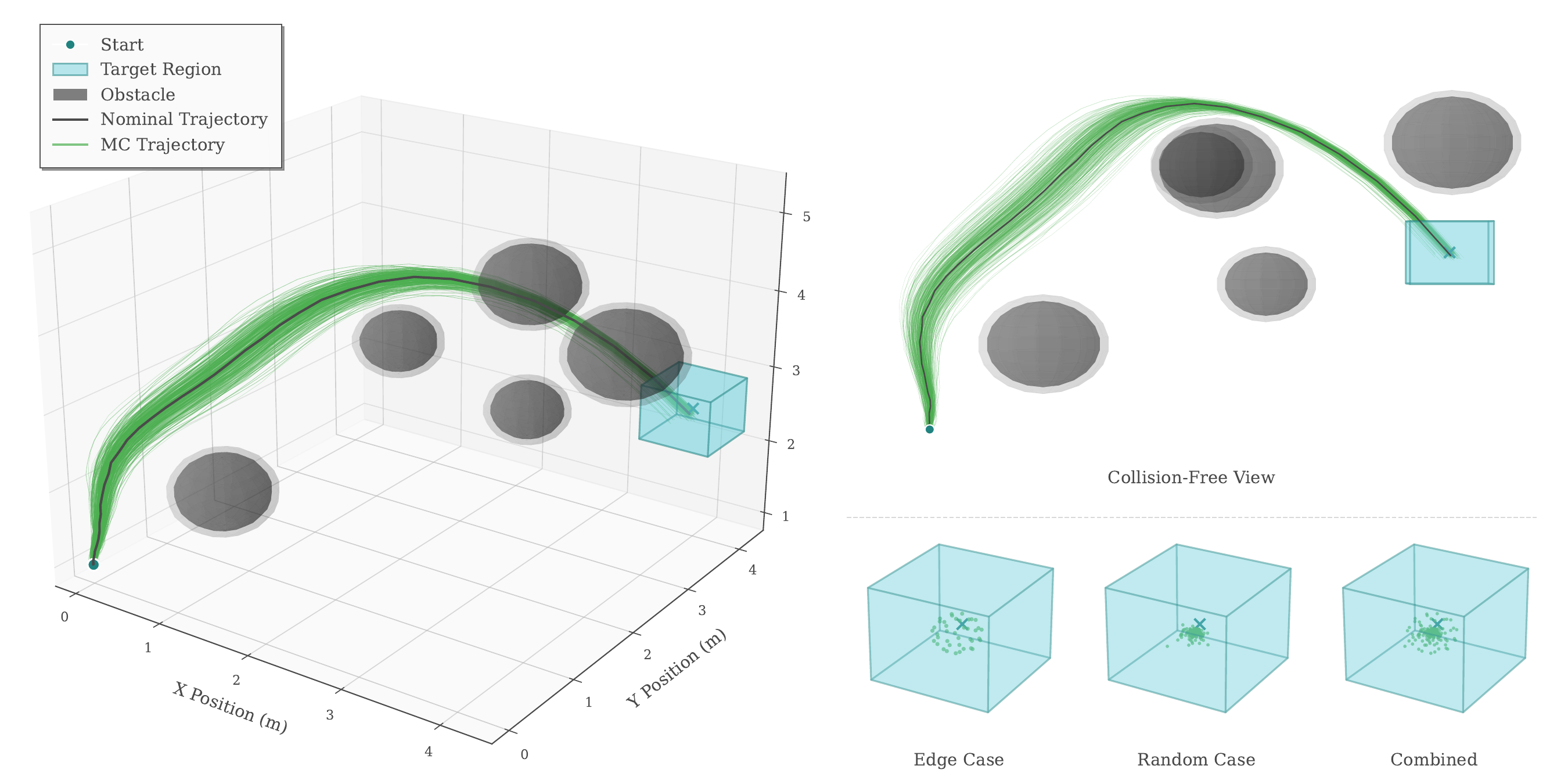}
        \caption{NRTO-FullADMM}
        \label{fig:quadcopter_nrto_fulladmm_demo}
    \end{subfigure}\hfill

    \caption{
    \textbf{Performance comparison on Quadcopter model with five obstacles:}
    We compare (a) NRTO-DR and (b) NRTO-FullADMM; both produce collision-free trajectories for 2,000 disturbance realizations. Insets report the distribution of terminal state realizations under random Monte Carlo, boundary edge-case, and combined disturbances.}
    \label{fig:exp_quadcopter}
\end{figure*}

\begin{table*}[t]
\centering
\scalebox{1.05}{
\begin{tabular}{@{}llll|ccc|ccc@{}}
\toprule
& & & & \multicolumn{3}{c|}{\textbf{Unicycle}} & \multicolumn{3}{c}{\textbf{Quadcopter}} \\
\cmidrule(lr){5-7}\cmidrule(lr){8-10}
\textbf{Sweep} & \textbf{Setting} & \textbf{Controls} & \textbf{Framework}
& \makecell{\textbf{NRTO}\\\textbf{Succ. (\%)}} 
& \makecell{\textbf{NRTO LE}\\\textbf{Succ. (\%)}} 
& \textbf{Time (s)}
& \makecell{\textbf{NRTO}\\\textbf{Succ. (\%)}} 
& \makecell{\textbf{NRTO LE}\\\textbf{Succ. (\%)}} 
& \textbf{Time (s)} \\
\midrule

\multirow{9}{*}{Horizon $T$}
& \multirow{3}{*}{$T=15$}
& \multirow{3}{*}{\makecell{$\tau=\tau_0$\\$Obs = O_0$}}
& NRTO        & \textbf{100}        & \textbf{100} & 12.013             & \textbf{100} & \textbf{100} & 952.603 \\
&  &  & NRTO-DR        & \textbf{100}        & \textbf{100} & \textbf{5.324}    & \textbf{100} & \textbf{100} & \underline{197.171} \\
&  &  & NRTO-FullADMM  & \underline{99}     & \textbf{100} & \underline{5.983} & \underline{99} & \textbf{100} & \textbf{43.543} \\
\cmidrule(lr){2-10}

& \multirow{3}{*}{$T=30$}
& \multirow{3}{*}{\makecell{$\tau=\tau_0$\\$Obs = O_0$}}
& NRTO        & \textbf{91}        & \textbf{100} & 32.137             & \underline{98} & \textbf{100} & 1327.390 \\
&  &  & NRTO-DR        & \textbf{91}        & \textbf{100} & \underline{7.328} & \underline{98} & \textbf{100} & \underline{218.259} \\
&  &  & NRTO-FullADMM  & \underline{90}     & \textbf{100} & \textbf{6.231}    & \textbf{99} & \textbf{100} & \textbf{98.585} \\
\cmidrule(lr){2-10}

& \multirow{3}{*}{$T=45$}
& \multirow{3}{*}{\makecell{$\tau=\tau_0$\\$Obs = O_0$}}
& NRTO        & \underline{82}        & \textbf{100} & 44.324 & \underline{97.5} & \textbf{100} & 1749.458 \\
&  &  & NRTO-DR        & \underline{82}        & \textbf{100} & \underline{7.847}             & 97 & \textbf{100} & \underline{238.313} \\
&  &  & NRTO-FullADMM  & \textbf{84}     & \textbf{100} & \textbf{6.387}    & \textbf{98} & \textbf{100} & \textbf{132.484} \\

\midrule

\multirow{9}{*}{Uncertainty $\tau$}
& \multirow{3}{*}{$\tau=0.01$}
& \multirow{3}{*}{\makecell{$T=T_0$\\$Obs = O_0$}}
& NRTO        & \textbf{100}       & \textbf{100} & 29.843            & \textbf{100} & \textbf{100} & 1132.324 \\
&  &  & NRTO-DR        & \textbf{100}       & \textbf{100} & \underline{7.671} & \underline{99} & \textbf{100} & \underline{65.454} \\
&  &  & NRTO-FullADMM  & \textbf{100}     & \textbf{100} & \textbf{6.988}    & \underline{99} & \textbf{100} & \textbf{19.452} \\
\cmidrule(lr){2-10}

& \multirow{3}{*}{$\tau=0.05$}
& \multirow{3}{*}{\makecell{$T=T_0$\\$Obs = O_0$}}
& NRTO       & \textbf{91}        & \textbf{100} & 31.543            & \textbf{99} & \textbf{100} & 1332.384 \\
&  &  & NRTO-DR        & \textbf{91}        & \textbf{100} & \underline{8.031}& \underline{98} & \textbf{100} & \underline{248.473} \\
&  &  & NRTO-FullADMM  & \underline{90}     & \textbf{100} & \textbf{6.593}    & \underline{98} & \textbf{100} & \textbf{97.482} \\
\cmidrule(lr){2-10}

& \multirow{3}{*}{$\tau=0.075$}
& \multirow{3}{*}{\makecell{$T=T_0$\\$Obs = O_0$}}
& NRTO        & \textbf{82}        & \textbf{100} & 33.314            & \underline{81} & \textbf{100} & 1371.342 \\
&  &  & NRTO-DR        & \textbf{82}        & \textbf{100} & \underline{7.623} & \underline{81} & \textbf{100} & \underline{739.546} \\
&  &  & NRTO-FullADMM  & \underline{81}     & \textbf{100} & \textbf{7.013}    & \textbf{82} & \textbf{100} & \textbf{211.483} \\

\midrule

\multirow{9}{*}{Obstacles}
& \multirow{3}{*}{$Obs=0$}
& \multirow{3}{*}{\makecell{$T=T_0$\\$\tau=\tau_0$}}
& NRTO        & \textbf{91}     & \textbf{100} & 31.153            & \textbf{99} & \textbf{100} & 1320.487 \\
&  &  & NRTO-DR        & \textbf{91}     & \textbf{100} & \underline{8.352}& \underline{98} & \textbf{100} & \underline{213.384} \\
&  &  & NRTO-FullADMM  & \underline{90}       & \textbf{100} & \textbf{6.324}   & \underline{98} & \textbf{100} & \textbf{98.345} \\
\cmidrule(lr){2-10}

& \multirow{3}{*}{$Obs=3$}
& \multirow{3}{*}{\makecell{$T=T_0$\\$\tau=\tau_0$}}
& NRTO        & \textbf{90}        & \textbf{100} & 898.342           & \underline{77} & \textbf{100} & 9142.582 \\
&  &  & NRTO-DR        & \underline{87}        & \textbf{100} & \underline{45.532}& \textbf{77.5} & \textbf{100} & \underline{6384.378} \\
&  &  & NRTO-FullADMM  & \underline{87}     & \textbf{100} & \textbf{43.483}   & \underline{77} & \textbf{100} & \textbf{121.392} \\
\cmidrule(lr){2-10}

& \multirow{3}{*}{$Obs=5$}
& \multirow{3}{*}{\makecell{$T=T_0$\\$\tau=\tau_0$}}
& NRTO        & \textbf{90}        & \textbf{100} & 30423.523         & \textbf{74} & \textbf{100} & 13374.375 \\
&  &  & NRTO-DR        & \underline{85}     & \textbf{100} & \underline{1934.314} & \underline{72} & \textbf{100} & \underline{9836.486} \\
&  &  & NRTO-FullADMM  & \underline{85}                 & \textbf{100}            & \textbf{217.932}  & 69 & \textbf{100} & \textbf{199.935} \\

\bottomrule
\end{tabular}
}
\caption{Unicycle and quadcopter dynamics. Three independent sweeps vary one factor while holding the others fixed at \((T_0,\tau_0,O_0)\). \textbf{Bold}: best; \underline{Underline}: second-best.}
\label{tab:nrto_sweep_unicycle_quadcopter}
\end{table*}

\section{Simulation} \label{Simulation Sec}
We evaluate cuNRTO, a suite comprising the GPU-accelerated versions of NRTO-DR and NRTO-FullADMM, on unicycle, quadcopter, and Franka manipulator tasks. We provide a comprehensive analysis of the impact of key system parameters, such as time horizon ($T$), uncertainty level ($\tau$), and the number of obstacles, on both constraint satisfaction and wall-clock time in comparison to the baseline NRTO. Furthermore, we highlight the effectiveness of the proposed architectures in addressing complex dynamical systems through a high-dimensional Franka manipulator task.

\subsection{Methodology}
Results were collected on a high-performance workstation with a 3.06~GHz 60-core Intel Xeon W-3500 CPU and an NVIDIA A100 80GB GPU, running Ubuntu 22.04 and CUDA 12.9. Code was compiled with g++~11.4.0. We use the same outer-loop parameters for all methods. 
The baseline NRTO solves subproblems using MOSEK \cite{mosek} interior-point optimizer with multi-threading enabled.
Solver-specific hyperparameters were tuned independently for best performance, and its complete list is provided in Section IV of SM.

The performance is evaluated based on the following:
\begin{itemize}
    \item \textbf{Constraint satisfaction:} Constraint satisfaction was verified using Monte Carlo sampling with 1{,}000 i.i.d. disturbance drawn uniformly from the interior of the uncertainty set, combined with 1,000 reproducible edge cases that probe the boundary of the uncertainty set via convex combinations of worst-case constraint directions. We set fixed seed to ensure the consistency of the sampling across all experiments. The reported satisfaction probability is the fraction of successful rollouts.
    \item \textbf{Wall-clock time:} We report the wall-clock time required for each solver to converge to a feasible solution. Runtimes were measured using \texttt{C++} \texttt{std::chrono::high\_resolution\_clock}.
\end{itemize}

\subsection{Performance Evaluation}
Trajectory comparisons between the proposed frameworks and the baseline NRTO for the unicycle and quadcopter models are presented in Fig.~\ref{fig:exp_unicycle} and Fig.~\ref{fig:exp_quadcopter}, respectively.
%
Further, the computational efficiency is demonstrated in Table~\ref{tab:nrto_sweep_unicycle_quadcopter}. 
We report three sweep studies over horizon length $T$, uncertainty level $\tau$, and the number of obstacles for both unicycle and quadcopter. We denote the nominal setting as $(T_0,\tau_0,O_0)=(30,0.05,0)$ and vary one parameter at a time. All experiments use time-step $\Delta k = $ 25~ms for both optimization and rollout.

\subsubsection{Horizon Sweep ($T$)}
At the nominal horizon $T_0=30$, NRTO-FullADMM reduces solve time from 32.137~s to 6.231~s for the unicycle, which is 5.16$\times$ faster, and from 1327.390~s to 98.585~s for the quadcopter, which is 13.46$\times$ faster.
At $T=45$, NRTO-FullADMM reaches 6.387~s for the unicycle and 132.484~s for the quadcopter, corresponding to 6.94$\times$ and 13.21$\times$ speedups over NRTO, respectively.

\subsubsection{Obstacle Sweep ($Obs$)}
As the number of obstacles increases, the number of robust constraints and associated SOC projections grow, and the GPU implementations benefit from the resulting parallelism.
At $Obs=5$, NRTO-FullADMM reduces solve time from 30423.523~s to 217.932~s for the unicycle, which is 139.60$\times$ faster, and from 13374.375~s to 199.935~s for the quadcopter, which is 66.89$\times$ faster.

\subsubsection{Uncertainty sweep ($\tau$)}
As $\tau$ increases, the NRTO success rate decreases, reflecting the increased conservativeness and difficulty of the robust constraints.
In contrast, when accounting for linearization error using NRTO-LE \cite{aabdulnonlinearcdc}, we observe 100\% constraint satisfaction across all reported $\tau$ values for both dynamics.
Across the sweep, NRTO-FullADMM consistently yields the lowest wall-clock times among the compared methods.

\subsection{Source of speedup}
For the unicycle task with five obstacles in Fig.~\ref{fig:exp_unicycle}, NRTO-DR spends most of its runtime on host--device synchronization and linear solves ($43.5\%$ and $32.6\%$), while SOC projections account for only $11.0\%$. A detailed runtime breakdown is provided in Section IV of the SM. By keeping the inner ADMM loop on-device, NRTO-FullADMM reduces synchronization overhead and increases average GPU utilization from $34.9\%$ to $86.5\%$.

\subsection{Solution quality}
The proposed reformulations achieve significant speedup without compromising solution quality. The visual variations in terminal-state distributions in {Fig. \ref{fig:exp_unicycle}-\ref{fig:exp_quadcopter}} and the linearization error differences in Table \ref{tab:nrto_sweep_unicycle_quadcopter} are due to convergence toward distinct local minima. These variations arise from the sensitivity of successive linearization towards initialization, hyperparameters, and solver tolerances, rather than a degradation of robust feasibility. 
Consequently, Table \ref{tab:nrto_sweep_unicycle_quadcopter} reveals no distinct pattern in linearization errors across the solvers. Furthermore, for the unicycle task (Fig.~\ref{fig:exp_unicycle}), the objective components $(J_{u},J_{kv})$ for NRTO, NRTO-DR, and NRTO-FullADMM are $(14.624,204.914)$, $(14.932,204.912)$, and $(13.117,207.425)$, respectively (refer to Table \ref{tab:solution_quality}), with less than $0.5\%$ difference across solvers.

\begin{figure*}[!h]
    \centering
    \includegraphics[width=\textwidth]{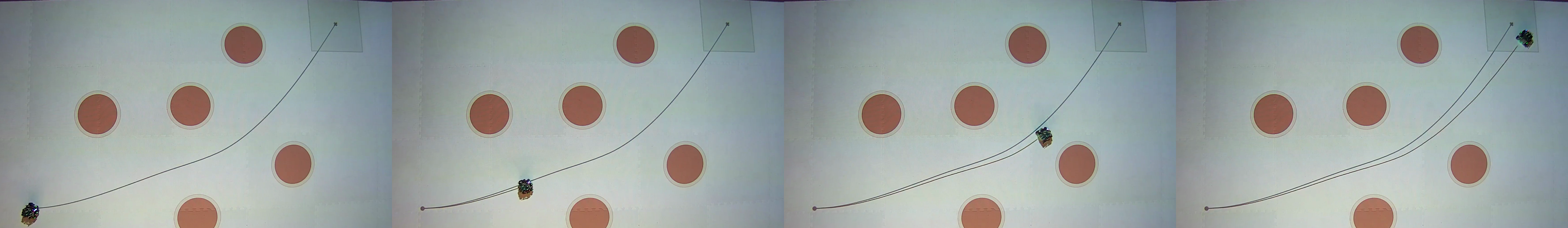}
    \caption{
    \textbf{Real-world Robotarium rollout with NRTO disturbance-feedback.}
    Four snapshots over time show the executed trajectory using the affine disturbance-feedback policy.
    The feedback gain significantly reduces accumulated tracking error and steers the robot into the target region despite real-world disturbances and actuation imperfections.
    }
    \label{fig:Real World Feedback Gain}
\end{figure*}
 
\begin{figure*}[t]
    \centering
    \includegraphics[width=\textwidth]{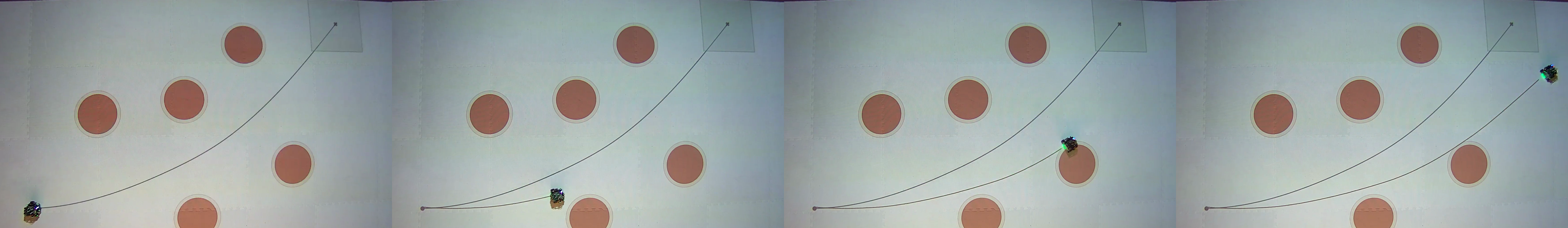}
    \caption{
    \textbf{Real-world Robotarium rollout with nominal control only.}
    Using only the nominal sequence leads to substantial drift from the planned path.
    The resulting tracking error accumulates over time, and the robot fails to reliably satisfy constraints.
    }
    \label{fig:Real World Nominal Only}
\end{figure*}

\begin{table}[t]
\centering
\scalebox{1.0}{
\begin{tabular}{@{}lccc@{}}
\toprule
\textbf{Objective Component}
& \textbf{NRTO}
& \textbf{NRTO-DR}
& \textbf{NRTO-FullADMM} \\
\midrule
$J_u = 0.05\|\bar{u}\|_2^2$
& \underline{14.624}
& 14.932
& \textbf{13.117} \\
$J_{K_v} = 0.1\|K_v\|_2^2$
& \underline{204.914}
& \textbf{204.912}
& 207.425 \\
Total
& \textbf{219.538}
& \underline{219.844}
& 220.542 \\
\bottomrule
\end{tabular}
}
\caption{Solution quality comparison for unicycle case (Fig. \ref{fig:exp_unicycle}). \textbf{Bold}: best; \underline{Underline}: second-best.}
\label{tab:solution_quality}
\end{table}




\subsection{Real-World Robotarium Experiment}
\label{sec:robotarium_exp}

To further demonstrate that the policy can compensate for real-world model mismatch, we deploy NRTO on a Robotarium unicycle robot. The disturbance set is estimated from open-loop rollout residuals, and the resulting NRTO policy applies the affine feedback $u_k=\bar{u}_k+K_k d_k$, where $d_k$ is estimated online from consecutive state-transition residuals. As shown in Fig.~\ref{fig:Real World Feedback Gain}--\ref{fig:Real World Nominal Only}, the policy substantially reduces accumulated tracking error and steers the robot into the target region, whereas executing the nominal control sequence alone leads to visible drift. Additional details on disturbance estimation are provided in Section V of SM.

\subsection{Franka Manipulator Experiment}
\begin{figure}[t]
    \centering
    \includegraphics[width=0.485\textwidth]{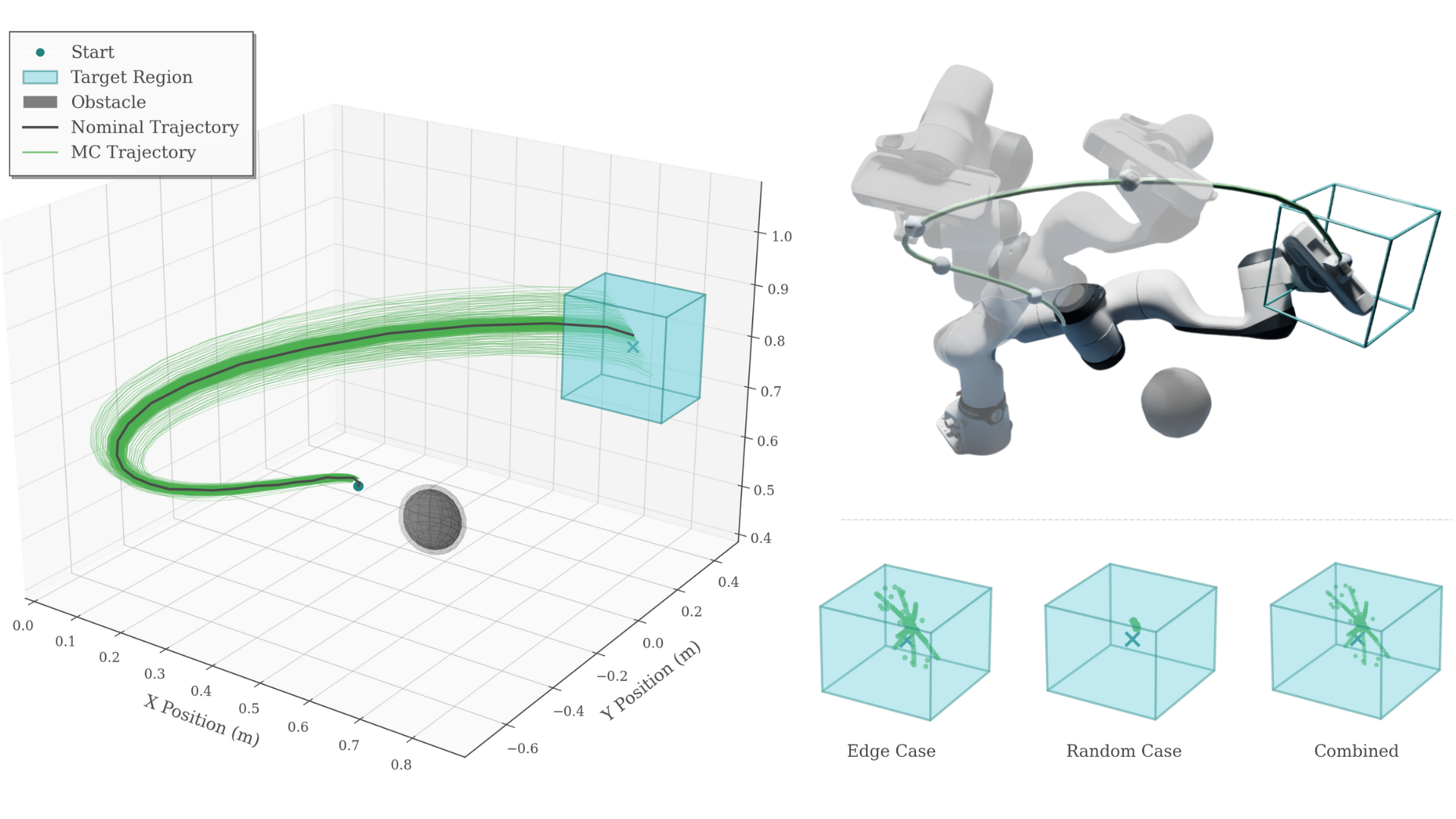}
    \caption{
    \textbf{NRTO-FullADMM on Franka Manipulator Task:}
    Left: end-effector trajectory with disturbance rollouts (green) toward the goal region (cyan) while avoiding obstacles (gray). Right: qualitative visualization of the motion in Isaac Sim \cite{NVIDIA_Isaac_Sim}.}
    \label{fig:franka_demo}
\end{figure}

We evaluate our framework on a 7-DOF Franka Emika Panda manipulator, a high-dimensional system with $n_{x}=14$ joint positions and velocities and $n_{u}=7$ joint torques, as shown in Fig.~\ref{fig:franka_demo}. The task is to drive the end-effector from a nominal configuration to a goal region while satisfying joint position limits, joint velocity limits, torque bounds, and collision avoidance constraints with respect to spherical obstacles in the workspace. Collision constraints are evaluated using Isaac Sim collision spheres. The feedback gain variables scale as $\mathcal{O}\left(T n_{u} n_{x}\right)$, yielding $\mathbf{K}_{k} \in \mathbb{R}^{7 \times 14}$ per timestep, and obstacle avoidance constraints are enforced at every knot point via the end-effector position obtained from forward kinematics. We report experiments across fixed horizon length $T = 30$, number of obstacles $n_{\mathrm{obs}} = 1$, and robustness level $\tau = 0.01$. 
As shown in Table~\ref{tab:nrto_franka_fixed}, NRTO-FullADMM achieves a 25.9$\times$ wall-clock speedup over NRTO on this Franka setting, indicating the scalability of NRTO-FullADMM.

\begin{table}[t]
\centering
\scalebox{1.0}{
\begin{tabular}{@{}lccc@{}}
\toprule
\textbf{Framework}
& \makecell{\textbf{NRTO}\\\textbf{Succ. (\%)}}
& \makecell{\textbf{NRTO-LE}\\\textbf{Succ. (\%)}}
& \textbf{Time (s)} \\
\midrule
NRTO           & \textbf{96} & \textbf{100} & 3948.523 \\
NRTO-DR        & \textbf{96} & \textbf{100} & \underline{1342.455} \\
NRTO-FullADMM  & \underline{95} & \textbf{100} & \textbf{152.384} \\
\bottomrule
\end{tabular}
}
\caption{Franka manipulator dynamics with fixed setting $(T,\tau,\mathrm{Obs})=(30,0.01,1)$. \textbf{best}; \underline{second-best}.}
\label{tab:nrto_franka_fixed}
\end{table}

\section{Conclusion}
We introduced cuNRTO, a GPU-accelerated implementation of nonlinear robust trajectory optimization (NRTO) \cite{aabdulnonlinearcdc}.
By exposing fine-grained parallelism in second-order cone (SOC) projections and reusing constant linear operators across inner iterations, cuNRTO enables two accelerated inner solvers, NRTO-DR and NRTO-FullADMM. 
Across unicycle, quadcopter, and Franka manipulator tasks, the proposed methods achieve substantial wall-clock speedups up to 139.6$\times$ over the baseline while maintaining robust constraint satisfaction.

In future work, we will extend cuNRTO to contact-rich manipulation and locomotion settings involving larger conic programs. We aim to further enhance computational speed by leveraging learning-to-optimize frameworks \cite{chen2022learning, tang2024learn}, ranging from learning-to-warmstart architectures \cite{banerjee2020learning, celestini2024transformer, zinage2025transformermpc} to deep-unfolding techniques \cite{saravanos2024deep, 11201755, oshin2025deep}. Another direction is to extend the framework's applicability to multi-agent swarms and to address heterogeneous forms of uncertainty \cite{Kotsalis_covariance_steering, DRO_ACC}.

\section*{Acknowledgments}
Arshiya Taj Abdul and Evangelos A. Theodorou are supported by the ARO Award \#W911NF2010151.

\newpage

\bibliographystyle{IEEEtran}
\bibliography{references}

\clearpage
\section*{Supplementary Material}
\setcounter{equation}{17}
\setcounter{table}{2}

\maketitle
\section{Details related to NRTO framework}
\label{Details related to NRTO framework}
\subsection{Tractable Linearized Problem}
The cost functions are given as
\begin{align}
\pazocal{Q}_{\hat{u}} (\delta \hat{\bu}) 
&=
\sum_{k=0}^{T-1} ( \hat{\bu}_k + \delta \hat{\bu}_k )^{\top}\, \vR_u^k\, ( \hat{\bu}_k + \delta \hat{\bu}_k),
\\
\tilde{\pazocal{Q}} (\bk_v)
&=
\sum_{k=0}^{T-1} \| \vR_K^k \vK_k \|_F^2
~=~
\frac{1}{2}\bk_v^{\top} \vQ_v \bk_v,
\\
\vQ_v
&:= 2\,\operatorname{blkdiag}\!\big( \vQ_{v,0},\ldots,\vQ_{v,T-1} \big),
\\
\vQ_{v,k}
&:= \vI_{n_x}\otimes (\vR_K^k)^{\top} \vR_K^k.
\end{align}
The function $F_u$ and the corresponding matrices involved can be referred from [18]. In this work, the constraint $g_j^{\text{trac}}(\vK, \bp) \leq 0$ from Problem 3 of [18] is rewritten in the form of $ \| \hat{\vA}_j \bk_v + \hat{\bb}_j \|_2 \leq p_j$. Since $\vS \succ 0$, there exists $\Psi \succ 0$ such that $\Psi \T \Psi = \vS^{-1}$. The matrices $\hat{\vA}_j$ and $\hat{\bb}_j$ can be given as follows 

\begin{align}
\text{Let}\quad
\bb_j
&:= \vF_u\T \nabla_x \bg_j(\hat{\bx}) \in \Rb^{T n_u},
\\
\bb_j
&= [\bb_{j,0};\ldots;\bb_{j,T-1}],
\qquad \bb_{j,k}\in\Rb^{n_u},
\\
\bar{\vA}_{j,k}
&:= \vI_{n_x}\otimes \bb_{j,k}\T,
\qquad \forall k\in\llbracket 0, T-1\rrbracket,
\\
\bar{\vA}_j
&:= \mathrm{blkdiag}\!\big(\bar{\vA}_{j,0},\ldots,\bar{\vA}_{j,T-1}\big),
\\
\hat{\vA}_j
&= \sqrt{\tau}\,\Psi\,\vGamma\T
\begin{bmatrix}
\bar{\vA}_j \\ \mathbf{0}_{n_x\times (T n_u n_x)}
\end{bmatrix},
\\
\hat{\bb}_j
&= \sqrt{\tau}\,\Psi\,\vGamma\T \vF_{\zeta}\T \nabla_x \bg_j(\hat{\bx}).
\end{align}
\noindent
We used $\vK_k\T \bb_{j,k} = (\vI_{n_x}\otimes \bb_{j,k}\T)\mathrm{vec}(\vK_k)$ for column-wise vectorization.

%
%
%
\section{Proofs related to Framework-1}
\label{Proofs related to Framework-1}

\subsection{Standard conic form}
\label{app:dr-standardform}

Define $\bchi := [\bk_v;\tilde{\bp}]\in\Rb^{ T n_u n_x +n_g}$.
The quadratic objective in (7a) can be written as $\tfrac{1}{2}\bchi^\rT \vP \bchi + \bq^\rT \bchi$ where
\begin{equation}
\begin{aligned}
\vP=\mathrm{blkdiag}(\vQ_v,\rho\vI_{n_g}),\quad
\bq=\begin{bmatrix}\mathbf{0}\\ -\rho(\bp^{l_{\text{in}} -1} + \blambda^{ l_{\text{in}} -1} / \rho)\end{bmatrix}.
\nonumber
\end{aligned}
\end{equation}

For each SOC constraint $j$, introduce a slack block $\bs_j=(t_j,\boldsymbol{\eta}_j)\in\Rb\times\Rb^{n_z}$ and a conic block $\calK_j$ given as 
\begin{equation}
\calK_j := \{(t,\boldsymbol{\eta}): \|\boldsymbol{\eta}\|_2\le t\}
\end{equation}
Impose the affine relations
\begin{equation}
\label{eq:app-affine-rel}
t_j=\tilde{p}_j,\qquad \boldsymbol{\eta}_j=\hat{\vA}_j\bk_v+\hat{\bb}_j,\qquad j\in\llbracket 1,n_g\rrbracket,
\end{equation}
and stack $\bs=[\bs_1;\ldots;\bs_{n_g}]\in\Rb^{m}$ with $m=n_g(1+n_z)$.
Then \eqref{eq:app-affine-rel} can be written as a single linear constraint
\begin{equation}
\label{eq:app-standard}
\vA\bchi+\bs=\mathbf{b},\qquad \bs\in\calK,
\end{equation}
where $\calK := \calK_1\times\cdots\times\calK_{n_g}$, and the matrices
$\vA$ and $\mathbf{b}$ are formed by stacking per-constraint blocks

\begin{equation}
\label{eq:app-A-block}
\begin{aligned}
\vA_j &:= 
\begin{bmatrix}
\mathbf{0}_{1\times (T n_u n_x)} & -\be_j^\rT\\
-\hat{\vA}_j & \mathbf{0}_{n_z\times n_g}
\end{bmatrix},\qquad
\mathbf{b}_j :=
\begin{bmatrix}
0\\
\hat{\bb}_j
\end{bmatrix},\\
\vA &=
\begin{bmatrix}
\vA_1\\ \vdots\\ \vA_{n_g}
\end{bmatrix},
\qquad
\mathbf{b} = [\mathbf{b}_1;\ldots;\mathbf{b}_{n_g}] .
\end{aligned}
\end{equation}

and $\be_j$ is the $j$-th standard basis vector in $\Rb^{n_g}$.

\subsection{Affine-set projection KKT system}
\label{app:dr-kkt-derivation}
The DR update (11a) involves projection onto affine-set and can be explicitly given as follows
%
\begin{equation}
\label{eq:app-Lproj}
\begin{aligned}
\min_{\bchi,\bs} ~
& \tfrac{1}{2}\bchi^\rT \vP \bchi + \bq^\rT \bchi
+ \tfrac{1}{2}\|\bchi - \tilde{\bchi}^{ l_{\text{dr}} - 1 } \|_{\vR_\chi}^2
+ \tfrac{1}{2}\|\bs - \tilde{\bs}^{l_{\text{dr}} - 1}  \|_{\vR_s}^2 \\
\text{s.t.}
& \quad \vA\bchi+\bs=\mathbf{b}. \nonumber
\end{aligned}
\end{equation}
with $\vR_\chi,\vR_s\succ 0$.
Introducing a multiplier $\by$ and eliminating $\bs$ via optimality gives

\begin{equation}
\label{eq:app-KKTfinal}
\begin{aligned}
\underbrace{\begin{bmatrix}
\vP+\vR_\chi & \vA^\rT\\
\vA & -\vR_s^{-1}
\end{bmatrix}}_{\vK_{\mathrm{KKT}}}
\begin{bmatrix} \bchi^{ l_{\text{dr}} } \\ \by^{ l_{\text{dr}}  } \end{bmatrix}
&=
\begin{bmatrix}
\vR_\chi \tilde{\bchi}^{ l_{\text{dr}} - 1 } - \bq \\
        \mathbf{b}- \tilde{\bs}^{ l_{\text{dr}} - 1 }
\end{bmatrix},\\
\bs^{ l_{\text{dr}}  } = \tilde{\bs}^{ l_{\text{dr}} - 1 } - \vR_s^{-1} \by^{ l_{\text{dr}} } \nonumber
\end{aligned}
\end{equation}

Within one SL iteration, $\vK_{\mathrm{KKT}}$ is constant, and only the RHS changes, enabling factorization reuse.

\subsection{Closed-form SOC projection}
\label{app:soc-proj}
The DR update (11c) involves $\text{Prox}_{\hat{\calJ}}(\bxi_{\text{ref}}^{l_{\text{dr}}})$.
Since $\hat{\calJ}(\bs)=\calI_{\calK}(\bs)$ depends only on $\bs$, we use the Euclidean proximal metric for the cone step:
\begin{equation}
\begin{aligned}
     \min_{ \hat{\bchi}, \hat{\bs}} ~~ & 
     \tfrac{1}{2}\|\hat{\bchi} - \bchi_{\text{ref}}^{ l_{\text{dr}} } \|_2^2 
     + \tfrac{1}{2}\| \hat{\bs} - \bs_{\text{ref}}^{l_{\text{dr}} }  \|_{2}^2 
     \\
     \text{s.t. } ~~ & \hat{\bs} \in \calK.
\end{aligned}
\end{equation}
Therefore,
\begin{equation}
    \text{Prox}_{\hat{\calJ}} (\bxi_{\text{ref}}^{ l_{\text{dr}} } )
    = \begin{bmatrix}
        \bchi_{\text{ref}}^{ l_{\text{dr}} } \\
        \Pi_{\calK} ( \bs_{\text{ref}}^{l_{\text{dr}} }  )
    \end{bmatrix}.
\end{equation}
where $\Pi_{\calK}$ is projection onto the conic set $\calK$. 
Since $\calK=\calK_1\times\cdots\times\calK_{n_g}$ is a product cone, $\Pi_{\calK}$ applies
\eqref{eq:app-soc-projection} independently to each block $\bs_j$.
Thus, we get $\Pi_{\calK} ( \bs_{\text{ref}}^{l_{\text{dr}} }  ) 
= [ \Pi_{\calK_1} ( \bs_{\text{ref},1}^{l_{\text{dr}} }  ) ; \Pi_{\calK_2} ( \bs_{\text{ref},2}^{l_{\text{dr}} }  ); \dots \Pi_{\calK_{n_g}} ( \bs_{\text{ref}, n_g}^{l_{\text{dr}} }  )  ]$. 
For each $\calK_{\mathrm{soc}}=\{(t,\boldsymbol{\eta}):\|\boldsymbol{\eta}\|_2\le t\}$ and input $\br=(r^t,\br^{\eta})$ with $a=\| \br^{\eta } \|_2$,
the Euclidean projection is
\begin{equation}
\label{eq:app-soc-projection}
\Pi_{\calK_{\mathrm{soc}}}(r^t,\br^\eta)=
\begin{cases}
(r^t,\br^\eta), & a \le r^t,\\
(0,\mathbf{0}), & a \le -r^t,\\
\left(\dfrac{r^t+a}{2},\ \dfrac{r^t+a}{2a}\,\br^\eta\right), & \text{otherwise.}
\end{cases}
\end{equation}
\section{Proofs related to Framework-2}
\label{Proofs related to Framework-2}
\subsection{AL of Problem (12):} The Augmented Lagrangian of the problem (12) is given as 
\begin{equation}
\begin{aligned}
    & \calL_{\rho}  ( \delta \hat{\bu}, \bk_v, \bnu,  \bp, \tilde{\bp}; \blambda) \\
    &
    =  \pazocal{H}_{p} ( \delta \hat{\bu}, \bp ) 
    + \tilde{\pazocal{Q}} (\bk_v) 
    + \calI_{\calK} (\bnu, \tilde{\bp})
    + \frac{\rho}{2} \| \bp - \tilde{\bp} + \blambda_{p} \|_2^2
    \\
    &~~~~~~~
    + \sum_{j = 1}^{n_g}
    \frac{\rho}{2} \| \hat{\vA}_j \bk_v + \hat{\bb}_j - \bnu_j 
    + \blambda_{\nu} \|_2^2
\end{aligned}
\end{equation}
where $\pazocal{H}_{p} ( \delta \hat{\bu}, \bp )  = \pazocal{Q}_{\hat{u}} (\delta \hat{\bu}) + \calI_{\calP} ( \delta \hat{\bu}, \bp )$, and $\calP$ and $\calK$ correspond to the constraints (12a) and (12b) respectively. The penalty parameter is $\rho$; and $\rho \blambda_p, \rho \blambda_{\nu}$ are the dual variables. 
\subsection{Derivation of ADMM update }
Considering $(\bnu, \tilde{\bp})$ are the first block, and $(\delta \hat{\bu}, \bp, \bk_v)$ as the second block of the ADMM, we have the block update steps given as follows 
\begin{align}
    & \{ \bnu, \tilde{\bp} \}^{ l_{\text{in}} }  \nonumber
    \\
    &~~ =
    \argmin_{ \bnu, \tilde{\bp}  } 
    \calL_{\rho}  ( \delta \hat{\bu}^{ l_{\text{in}} -1 }, \bk_v^{l_{\text{in}}-1} , \bnu,  \bp^{l_{\text{in}} -1 }, \tilde{\bp}; \blambda^{ l_{\text{in}} -1 } )
    \nonumber
    \\
    &  \{ \delta \hat{\bu}, \bp, \bk_v \}^{ ^{ l_{\text{in}} } }
    =
    \argmin_{\delta \hat{\bu}, \bp, \bk_v} 
    \calL_{\rho}  ( \delta \hat{\bu}, \bk_v , \bnu^{ l_{\text{in}}  },  \bp, \tilde{\bp}^{ l_{\text{in}}  }; \blambda^{ l_{\text{in}} -1 } )
    \nonumber
\end{align}
and the dual update steps as 
\begin{align}
    & \blambda_p^{l_{\text{in}}} =
    \blambda_p^{l_{\text{in}} -1 }
    + (\bp^{l_{\text{in}}} - \tilde{\bp}^{l_{\text{in}}}) \nonumber
    \\
    & \blambda_{\nu}^{l_{\text{in}}} =
    \blambda_{\nu}^{l_{\text{in}} -1 }
    + ( \hat{\vA}_j \bk_v^{l_{\text{in}}} + \hat{\bb}_j - \bnu^{ l_{\text{in}} }_j ) 
    \nonumber
\end{align}
Let us now simplify the block update steps. 
\subsubsection{Derivation of Block-1 update (13)}
This involves solving the following 
\begin{align}
    & \min_{ \bnu, \tilde{\bp}  } ~~~~
    \sum_{j = 1}^{n_g}
    \frac{\rho}{2} \| \hat{\vA}_j \bk_v^{ l_{\text{in}} -1 } + \hat{\bb}_j - \bnu_j 
    + \blambda_{\nu, j}^{ l_{\text{in}} -1 } \|_2^2 \nonumber
    \\
    &~~~~~~~~~~~~~~~~~
    + \frac{\rho}{2} \| \bp^{ l_{\text{in}} -1  } - \tilde{\bp} + \blambda_{p}^{l_{\text{in}} -1 } \|_2^2 \nonumber
    \\
    & ~~ \text{s.t. }
    \| \bnu_j \|_2 \leq \tilde{p}_j 
    ~~
    j \in \llbracket 1, n_g \rrbracket 
\end{align}
Note that the above update can be decoupled with respect to the variables $\{ \bnu_j, \tilde{p}_j \}_{j = 1}^{n_g}$. Therefore, we can update the variables in parallel as follows 
\begin{align}
   (\bnu_j^{l_{\text{in}} }, \tilde{p}_j^{ l_{\text{in}} })  
    & =
    \argmin_{ \bnu_j , \tilde{\bp}_j  } 
     \| \hat{\vA}_j \bk_v^{ l_{\text{in}} -1 } + \hat{\bb}_j - \bnu_j 
    + \blambda_{\nu, j}^{ l_{\text{in}} -1 } \|_2^2 \nonumber
    \\
    &~~~~~~~~~~~~~~~~~
    + \| \bp_j^{ l_{\text{in}} -1 } - \tilde{\bp}_j + \blambda_{p, j}^{ l_{\text{in}} -1 } \|_2^2 \nonumber
    \\
    \text{s.t. } & 
    \| \bnu_j \|_2 \leq \tilde{p}_j 
    ~~
    j \in \llbracket 1, n_g \rrbracket   
\end{align}
The above update has a closed form solution and can be given as in (13).
\subsubsection{Derivation of Block-2 updates ((14) and (15))}
This update involves solving the following 
\begin{align}
    &
    \min_{\delta \hat{\bu}, \bp, \bk_v} 
    \pazocal{Q}_{\hat{u}} (\delta \hat{\bu}) 
    + \tilde{\pazocal{Q}} (\bk_v) 
    + \frac{\rho}{2} \| \bp - \tilde{\bp}^{ l_{\text{in}} } + \blambda_{p}^{ l_{\text{in}} -1 } \|_2^2
    \nonumber
    \\
    &~~~~~~~~~~
    + \sum_{j = 1}^{n_g}
    \frac{\rho}{2} \| \hat{\vA}_j \bk_v + \hat{\bb}_j - \bnu^{ l_{\text{in}} }_j 
    + \blambda_{\nu}^{ l_{\text{in}} -1 } \|_2^2
    \nonumber
    \\
    & \text{s.t. } ~~ (12a) 
\end{align}
The above update can be decoupled with respect to the variables $(\delta \hat{\bu}, \bp)$ and $\bk_v$ as follows
\begin{align}
    & (\delta \hat{\bu}^{l_{\text{in}}}, \bp^{ l_{\text{in} }})
    = 
    \argmin_{\delta \hat{\bu}, \bp } \pazocal{Q}_{\hat{u}} (\delta \hat{\bu})
    + \frac{\rho}{2} \| \bp - \tilde{\bp}^{l_{\text{in}}} + \blambda_p^{l_{\text{in}} - 1} \|_2^2 \nonumber \\
    & ~~~~~~~~~~~~~~~~~~~~~~
    \text{s.t. }  (12a) 
    \\
    & \bk_v^{l_{\text{in}}} 
    = \argmin_{ \bk_v }
     \sum_{j = 1}^{n_g}
    \frac{\rho}{2} \| \hat{\vA}_j \bk_v + \hat{\bb}_j - \bnu^{  l_{\text{in}} }_j 
    + \blambda_{\nu}^{ l_{\text{in}} -1 } \|_2^2  \nonumber
    \\
    & ~~~~~~~~~~~~~~~~~~~~~~~
    + \frac{1}{2} \bk_v \T \vQ_v \bk_v 
\end{align}
A closed form solution can be obtained for the $\bk_v$ update. Let us start by writing the first order optimality conditions as follows 
\begin{align}
    & \sum_{j = 1}^{n_g} 
    \rho \hat{\vA}_j\T ( \hat{\vA}_j \bk_v^{ l_{\text{in}} } + \hat{\bb}_j - \bnu^{ l_{\text{in}} }_j 
    + \blambda_{\nu}^{ l_{\text{in}} -1 } )
    + \vQ_v \bk_v^{ l_{\text{in}} } = 0 \nonumber
\end{align}
using the dual update (17), we get the following
\begin{align}
    & \sum_{j = 1}^{n_g} 
    \rho \hat{\vA}_j\T ( \hat{\vA}_j \bk_v^{ l_{\text{in}} } + \hat{\bb}_j - \bnu^{ l_{\text{in}} }_j 
    + \blambda_{\nu}^{ l_{\text{in}} -1 } )
    + \vQ_v \bk_v^{ l_{\text{in}} } = 0 \nonumber
    \\
    \implies 
    &
    \sum_{j = 1}^{n_g} 
    \rho \hat{\vA}_j\T ( \blambda_{\nu}^{ l_{\text{in}} } )
    + \vQ_v \bk_v^{ l_{\text{in}} } = 0  
\end{align}
Assuming we choose $ \blambda_{\nu}^{0}, \bk_v^0 $ such that $ \sum_{j = 1}^{n_g} \rho \hat{\vA}_j\T ( \blambda_{\nu}^{ 0 } ) + \vQ_v \bk_v^{ 0 } = 0 $, we get the following from the above two equations 
\begin{align}
    & \sum_{j = 1}^{n_g} 
    \rho \hat{\vA}_j\T ( \hat{\vA}_j \bk_v^{ l_{\text{in}} } + \hat{\bb}_j - \bnu^{l_{\text{in}} }_j )
    + \vQ_v ( \bk_v^{ l_{\text{in}} } - \bk_v^{ l_{\text{in}} -1 } ) = 0
    \nonumber
    \\
    & \implies 
    \bigg( \vQ_v + \sum_{j = 1}^{n_g} \rho \hat{\vA}_j\T \hat{\vA}_j \bigg) \bk_v^{ l_{\text{in}} }  \nonumber
    \\[-0.2cm]
    & ~~~~~~~~~~~~~~~~~~ =
    \vQ_v \bk_v^{ l_{\text{in}} -1 } 
    + \sum_{j = 1}^{n_g} \rho \hat{\vA}_j\T (\bnu^{  l_{\text{in}} }_j - \bb_j ) 
\end{align}
Since $\vQ_v \succ 0$, we can write
\begin{equation}
\bk_v^{l_{\text{in}}} 
    = \bq + \calM \bk_v^{l_{\text{in}} - 1}  
    + \bar{\calM} \bnu^{ l_{\text{in}}}
\end{equation}
where 
\begin{align}
    & \vM = \bigg( \vQ_v + \sum_{j = 1}^{n_g} \rho \hat{\vA}_j\T \hat{\vA}_j \bigg)^{-1}, \nonumber
    \\
    & 
    \bq = - \sum_{j = 1}^{n_g} \rho \vM \hat{\vA}_j\T \hat{\bb}_j,   
    ~~ \calM = \vM \vQ_v, 
    \\
    &
    \bar{\calM} = \begin{bmatrix}
        \bar{\calM}_1 & \bar{\calM}_2 & \dots & \bar{\calM}_{n_g}
    \end{bmatrix}, 
    ~~ \bar{\calM}_j = \rho \vM \hat{\vA}_j\T \nonumber
\end{align}

 

\begin{table*}[h]
\centering
\scriptsize
\setlength{\tabcolsep}{4pt}
\begin{tabular}{@{}llp{0.8\textwidth}@{}}
\toprule
\textbf{Dynamics} & \textbf{$(n_x,n_u)$} & \textbf{Parameters and limits} \\
\midrule
Unicycle & $(3,2)$ &
\makecell[l]{
State $\mathbf{x}=[x,~y,~\theta]^\top$, control $\mathbf{u}=[v,~\omega]^\top$. \\
$(v_{\max},\omega_{\max})\in\{(3.0,1.5),(5,3)\}$.
} \\
\midrule
Quadcopter & $(12,4)$ &
\makecell[l]{
State $\mathbf{x}=[\mathbf{p}^\top,~\mathbf{v}^\top,~\phi,~\theta,~\psi,~\boldsymbol{\omega}^\top]^\top$, control $\mathbf{u}=[f,~\boldsymbol{\tau}^\top]^\top$. \\
$m=1.0$, $g=9.81$, $J=\mathrm{diag}(0.02,0.02,0.04)$. \\
$f\in[0,15]$, $\boldsymbol{\tau}\in[-0.5,0.5]^3$.
} \\
\midrule
Franka & $(14,7)$ &
\makecell[l]{
State $\mathbf{x}=[\mathbf{q}^\top,~\dot{\mathbf{q}}^\top]^\top$, control $\mathbf{u}=\boldsymbol{\tau}$. \\
$\mathbf{q}_{\min}=[-2.9007,-1.8361,-2.9007,-3.0770,-2.8763,0.4398,-3.0508]^\top$. \\
$\mathbf{q}_{\max}=[2.9007,1.8361,2.9007,-0.1169,2.8763,4.6216,3.0508]^\top$. \\
$\dot{\mathbf{q}}_{\max}=[2.62,2.62,2.62,2.62,5.26,4.18,5.26]^\top$,~~$\boldsymbol{\tau}_{\max}=[87,87,87,87,12,12,12]^\top$. \\
$\mathbf{d}=[0.5,\ldots,0.5]^\top$,~~$\mathbf{I}=[1,\ldots,1]^\top$.
} \\
\bottomrule
\end{tabular}
\caption{Dynamics dimensions and model parameters.}
\label{tab:sim_dynamics_params}
\end{table*}

\section{Simulation Data}
\label{Simulation Data}
\subsection{Dynamical Models}
\paragraph{Unicycle Model}
We use a planar unicycle with state $\mathbf{x}_k = [x_k,~y_k,~\theta_k]^\top$ and control $\mathbf{u}_k = [v_k,~\omega_k]^\top$. With forward-Euler discretization,
\begin{align}
    x_{k+1} &= x_k + v_k \cos(\theta_k)\Delta t, \\
    y_{k+1} &= y_k + v_k \sin(\theta_k)\Delta t, \\
    \theta_{k+1} &= \theta_k + \omega_k \Delta t,
\end{align}
with element-wise input saturation $|v_k|\le v_{\max}$ and $|\omega_k|\le \omega_{\max}$.

\paragraph{Quadcopter Model}
We use a 12D rigid-body quadcopter model with state
$\mathbf{x}_k = [\mathbf{p}_k^\top,~\mathbf{v}_k^\top,~\phi_k,~\theta_k,~\psi_k,~\boldsymbol{\omega}_k^\top]^\top$, where $\mathbf{p}=[x,y,z]^\top$, $\mathbf{v}=[v_x,v_y,v_z]^\top$, $(\phi,\theta,\psi)$ are roll--pitch--yaw angles, and $\boldsymbol{\omega}=[p,q,r]^\top$ are body rates. The control is $\mathbf{u}_k = [f_k,~\boldsymbol{\tau}_k^\top]^\top$, where $f$ is total thrust along the body $+z$ axis and $\boldsymbol{\tau}=[\tau_x,\tau_y,\tau_z]^\top$ are body torques. The continuous-time dynamics are
\begin{align}
    \dot{\mathbf{p}} &= \mathbf{v}, \\
    \dot{\mathbf{v}} &= -g \mathbf{e}_3 + \frac{f}{m} R(\phi,\theta,\psi)\mathbf{e}_3, \\
    \dot{\boldsymbol{\omega}} &= J^{-1}\!\left(\boldsymbol{\tau} - \boldsymbol{\omega} \times (J\boldsymbol{\omega})\right), \\
    \begin{bmatrix}\dot{\phi}\\\dot{\theta}\\\dot{\psi}\end{bmatrix}
    &= T(\phi,\theta)\boldsymbol{\omega},
\end{align}
where $\mathbf{e}_3=[0,0,1]^\top$, $R(\phi,\theta,\psi) = R_z(\psi)R_y(\theta)R_x(\phi)$, and
\begin{equation}
T(\phi,\theta)=
\begin{bmatrix}
1 & \sin\phi\tan\theta & \cos\phi\tan\theta \\
0 & \cos\phi & -\sin\phi \\
0 & \sin\phi/\cos\theta & \cos\phi/\cos\theta
\end{bmatrix}.
\end{equation}
We discretize with forward Euler as $\mathbf{x}_{k+1} = \mathbf{x}_k + \Delta t\,\dot{\mathbf{x}}_k$. Inputs are saturated element-wise to $f\in[f_{\min},f_{\max}]$ and $\boldsymbol{\tau}\in[\boldsymbol{\tau}_{\min},\boldsymbol{\tau}_{\max}]$.

\paragraph{Franka Manipulator Model}
We use a 7-DOF Franka Emika Panda joint-space model with state $\mathbf{x}_k=[\mathbf{q}_k^\top,~\mathbf{dq}_k^\top]^\top\in\mathbb{R}^{14}$ and control $\mathbf{u}_k=\boldsymbol{\tau}_k\in\mathbb{R}^{7}$, where $\mathbf{dq}=\dot{\mathbf{q}}$. The dynamics are
\begin{align}
    \dot{\mathbf{q}} &= \mathbf{dq}, \\
    \ddot{q}_i &= \frac{\tau_i - d_i\,dq_i}{I_i}, \qquad i=1,\ldots,7,
\end{align}
with diagonal damping $d_i$ and inertia $I_i$. The discrete-time update uses forward Euler with post-update clamping:
\begin{align}
    \mathbf{dq}_{k+1} &= \mathrm{clip}\!\left(\mathbf{dq}_k + \ddot{\mathbf{q}}_k\Delta t,~-\mathbf{dq}_{\max},~\mathbf{dq}_{\max}\right), \\
    \mathbf{q}_{k+1} &= \mathrm{clip}\!\left(\mathbf{q}_k + \mathbf{dq}_{k+1}\Delta t,~\mathbf{q}_{\min},~\mathbf{q}_{\max}\right),
\end{align}
and torque saturation $|\tau_i|\le \tau_{i,\max}$. End-effector position, orientation constraints, and obstacle distances are evaluated using forward kinematics.

\subsection{Experiment and Solver Parameters}
All numeric values and hyperparameters are shown in Tables~\ref{tab:sim_dynamics_params}--VIII.

\begin{table}[t]
\centering
\label{tab:hparams_uni_outer}
{\scriptsize
\setlength{\tabcolsep}{3pt}
\renewcommand{\arraystretch}{1.05}
\begin{tabularx}{\columnwidth}{@{}lY@{}}
\toprule
\textbf{Method} & \textbf{Outer loop} \\
\midrule
NRTO &
\makecell[l]{\textit{Outer Loop:} $N_{\text{Outer Loop}}=200$ \\
\textit{Trust region:} $r_0\in\{1.5,2\}$, $r_{\min}=0.001$ \\
\textit{Penalty:} $\rho_0=40$, $\rho_{\max}\in\{80,180\}$ \\
\textit{Acceptance:} $(\alpha,\beta,\eta_1,\eta_2)=(\{0.8,0.9\},\{1.15,1.5\},5,4)$ \\
\textit{Termination:} $(\epsilon_u,\epsilon_p)=(\{0.001,0.075\},\{10^{-4},0.01\})$ \\
\textit{Weights:} $w_p\in\{1,10\}$ \\
\textit{ADMM:} iters $=40$, $c_\epsilon=0.01$} \\
\midrule
NRTO-DR & \makecell[l]{same as NRTO} \\
\midrule
NRTO-FullADMM &
\makecell[l]{\textit{Outer Loop:} $N_{\text{Outer Loop}}=200$ \\
\textit{Trust region:} $r_0\in\{1.5,2\}$, $r_{\min}=0.001$ \\
\textit{Penalty:} $\rho_0=40$, $\rho_{\max}\in\{180,240,340\}$ \\
\textit{Acceptance:} $(\alpha,\beta,\eta_1,\eta_2)=(\{0.8,0.9\},\{1.15,1.5\},5,4)$ \\
\textit{Termination:} $(\epsilon_u,\epsilon_p)=(\{0.001,0.075\},\{10^{-4},0.05\})$ \\
\textit{Weights:} $w_p\in\{1,100,500\}$ \\
\textit{ADMM:} iters $=40$, $c_\epsilon=0.01$} \\
\bottomrule
\end{tabularx}}
\caption{Unicycle: outer-loop hyperparameters.}
\end{table}

\begin{table}[h]
\centering
\label{tab:hparams_uni_solver}
{\scriptsize
\setlength{\tabcolsep}{3pt}
\renewcommand{\arraystretch}{1.05}
\begin{tabularx}{\columnwidth}{@{}lY@{}}
\toprule
\textbf{Method} & \textbf{Solver-specific hyperparameters} \\
\midrule
NRTO &
\makecell[l]{MOSEK, CPU.} \\
\midrule
NRTO-DR &
\makecell[l]{$(\epsilon_{\mathrm{res}})=(10^{-4})$ \\
max iters: $100$, warm start: true.} \\
\midrule
NRTO-FullADMM &
\makecell[l]{\textit{FullADMM:} 
$\rho = 10$;
iters $=40$;\\ $\epsilon_p\in\{0.001,0.005\}$ \\
\textit{KV:} pcg; iters $\in\{100,1000\}$; tol $\in\{0.01,0.05\}$.} \\
\bottomrule
\end{tabularx}}
\caption{Unicycle: solver-specific hyperparameters.}
\end{table}

\begin{table}[h]
\centering
\label{tab:hparams_quad_outer}
{\scriptsize
\setlength{\tabcolsep}{3pt}
\renewcommand{\arraystretch}{1.05}
\begin{tabularx}{\columnwidth}{@{}lY@{}}
\toprule
\textbf{Method} & \textbf{Outer loop} \\
\midrule
NRTO &
\makecell[l]{\textit{Outer Loop:} $N_{\text{Outer Loop}}=200$ \\
\textit{Trust region:} $r_0\in\{1.5,2.5\}$, $r_{\min}=0.001$ \\
\textit{Penalty:} $\rho_0=40$, $\rho_{\max}=180$ \\
\textit{Acceptance:} $(\alpha,\beta,\eta_1,\eta_2)=(0.8,1.5,5,4)$ \\
\textit{Termination:} $(\epsilon_u,\epsilon_p)=(0.01,0.05)$ \\
\textit{Weights:} $w_p=10$ \\
\textit{ADMM:} iters $=40$, $c_\epsilon=0.01$} \\
\midrule
NRTO-DR &
\makecell[l]{\textit{Outer Loop:} $N_{\text{Outer Loop}}=200$ \\
\textit{Trust region:} $r_0\in\{1.5,2.5\}$, $r_{\min}=0.001$ \\
\textit{Acceptance:} $(\alpha,\beta,\eta_1,\eta_2)=(0.8,1.5,5,4)$ \\
\textit{Termination:} $(\epsilon_u,\epsilon_p)=(\{0.01,0.05\},0.01)$ \\
\textit{Weights:} $w_p\in\{10,100\}$ \\
\textit{ADMM:} iters $=40$, $c_\epsilon=0.01$} \\
\midrule
NRTO-FullADMM &
\makecell[l]{\textit{Outer Loop:} $N_{\text{Outer Loop}}=200$ \\
\textit{Trust region:} $r_0\in\{1.5,2.5\}$, $r_{\min}=0.001$ \\
\textit{Penalty:} $\rho_0=40$, $\rho_{\max}\in\{180,320\}$ \\
\textit{Acceptance:} $(\alpha,\beta,\eta_1,\eta_2)=(0.8,1.5,5,4)$ \\
\textit{Termination:} $(\epsilon_u,\epsilon_p)=(\{0.01,0.05\},\{0.01,0.05\})$ \\
\textit{Weights:} $w_p\in\{10,100\}$ \\
\textit{ADMM:} iters $=40$, $c_\epsilon=0.01$} \\
\bottomrule
\end{tabularx}}
\caption{Quadcopter: outer-loop hyperparameters.}
\end{table}

\begin{table}[h]
\centering
\label{tab:hparams_quad_solver}
{\scriptsize
\setlength{\tabcolsep}{3pt}
\renewcommand{\arraystretch}{1.05}
\begin{tabularx}{\columnwidth}{@{}lY@{}}
\toprule
\textbf{Method} & \textbf{Solver-specific hyperparameters} \\
\midrule
NRTO &
\makecell[l]{MOSEK, CPU} \\
\midrule
NRTO-DR &
\makecell[l]{$(\epsilon_{\mathrm{rel}})=(10^{-4})$ \\
max iters: $100$, warm start: true.} \\
\midrule
NRTO-FullADMM &
\makecell[l]{\textit{FullADMM:} 
$\rho = 10$;
iters $=40$; $\epsilon_p=0.001$ \\
\textit{KV:} pcg; iters $=100$; tol $=0.05$} \\
\bottomrule
\end{tabularx}}
\caption{Quadcopter: solver-specific hyperparameters.}
\end{table}

\begin{table}[t]
\centering
\label{tab:hparams_franka}
{\scriptsize
\setlength{\tabcolsep}{3pt}
\renewcommand{\arraystretch}{1.05}
\begin{tabularx}{\columnwidth}{@{}lY@{}}
\toprule
\textbf{Category} & \textbf{Hyperparameters} \\
\midrule
Outer &
\makecell[l]{\textit{Outer Loop:} $N_{\text{Outer Loop}}=80$ \\
\textit{Trust region:} $r_0=2$, $r_{\min}=0.001$ \\
\textit{Penalty:} $\rho_0=40$, $\rho_{\max}=80$ \\
\textit{Acceptance:} $(\alpha,\beta,\eta_1,\eta_2)=(0.8,1.5,5,4)$ \\
\textit{Termination:} $(\epsilon_u,\epsilon_p)=(0.0015,0.005)$ \\
\textit{Weights:} $w_p=250$ \\
\textit{ADMM:} iters $=80$, $c_\epsilon=0.01$} \\
\midrule
Solver-specific &
\makecell[l]{\textit{FullADMM:} $(\rho_s,\rho_p)=(10,10)$; iters $=50$; $\epsilon_p=0.001$ \\
\textit{KV:} pcg; iters $=100$; tol $=0.05$} \\
\bottomrule
\end{tabularx}}
\caption{Franka: outer-loop and solver-specific hyperparameters.}
\end{table}

\subsection{Runtime Breakdown and GPU Utilization}
\label{app:runtime_breakdown}

To better understand the source of the observed speedup, we profile the unicycle task with five obstacles, which is the setting where NRTO-FullADMM achieves the largest wall-clock improvement. As shown in Table~\ref{tab:runtime_breakdown}, NRTO-DR reduces the cost of the original SOCP solve but remains limited by host--device synchronization and sparse linear solves. In contrast, NRTO-FullADMM keeps the inner ADMM loop on-device, substantially reducing synchronization overhead and increasing average GPU utilization from $34.9\%$ to $86.5\%$. This indicates that the speedup is not due only to parallel SOC projections, but also to the algorithmic reformulation that removes the nested DR layer and reduces CPU--GPU communication.

\begin{table}[t]
\centering
\scriptsize
\setlength{\tabcolsep}{5pt}
\renewcommand{\arraystretch}{1.08}
\begin{tabular}{@{}lccc@{}}
\toprule
\textbf{Metric} & \textbf{NRTO} & \textbf{NRTO-DR} & \textbf{NRTO-FullADMM} \\
\midrule
\multicolumn{4}{@{}l}{\textit{Overall}} \\
Wall-clock time (s) & 30423.523 & 1934.314 & 217.932 \\
Successive linearization iterations & 56 & 71 & 197 \\
Average GPU busy (\texttt{nvidia-smi}) & -- & $34.9\%$ & $86.5\%$ \\
\midrule
\multicolumn{4}{@{}l}{\textit{Runtime share (\% of wall-clock)}} \\
Subproblem 1 total & $97.4$ & $87.1$ & -- \\
Conic solve + cuDSS (GPU) & -- & $87.1$ & -- \\
\quad host--device synchronization & -- & $43.5$ & -- \\
\quad linear system solve & -- & $32.6$ & -- \\
\quad parallel SOC projection & -- & $11.0$ & -- \\
Subproblem 2 total & $2.2$ & $12.9$ & -- \\
On-device inner ADMM loop & -- & -- & $95.7$ \\
\quad on-device kernels & -- & -- & $92.6$ \\
\quad host--device synchronization & -- & -- & $3.1$ \\
SL overhead & $0.4$ & $<0.1$ & $1.2$ \\
\bottomrule
\end{tabular}
\caption{Runtime breakdown and GPU utilization for the unicycle task with five obstacles.}
\label{tab:runtime_breakdown}
\end{table}

\section{Real-World Robotarium Experiment}
\label{app:robotarium}
\subsection{Platform and timing}
We consider a single Robotarium unicycle robot with state $x=[p_x,\,p_y,\,\theta]^\top$ and control $u=[v,\,\omega]^\top$.
The Robotarium simulator runs at a fixed internal timestep of $\Delta t_{\text{robot}}\approx 0.033$\,s. To keep the trajectory optimization tractable, we solve NTO/NRTO at a coarser timestep $\Delta t_{\text{solver}}=0.33$\,s. During deployment, we consider
each solver command is held for $M=\mathrm{round}(\Delta t_{\text{solver}}/\Delta t_{\text{robot}})$ Robotarium steps.

\subsection{Open-loop data collection (NTO)}
To estimate real-world disturbances without conflating them with a tracking controller, we first deploy an open-loop nominal trajectory produced by NTO.
At every Robotarium timestep, we log the measured robot pose $x_i^{\text{meas}}$ and the applied control $u_i^{\text{applied}}$.
We also record the commanded start pose $x_0^{\text{cmd}}$ and the measured start pose $x_0^{\text{meas}}$ for each rollout.

\subsection{Additive disturbance residuals}
We model the closed-loop rollout as the nominal dynamics plus an additive disturbance,
\begin{equation}
x_{k+1} = f(x_k, u_k, \Delta t_{\text{solver}}) + d_{k+1},
\end{equation}
On the solver grid we define:
\begin{align}
d_0 &:= x_0^{\text{meas}} - x_0^{\text{cmd}}, \\
d_{k+1} &:= x_{k+1}^{\text{meas}} - f(x_k^{\text{meas}}, u_k^{\text{applied}}, \Delta t_{\text{solver}}),
\quad k=0,\ldots,T-1,
\end{align}
where $x_k^{\text{meas}}$ and $u_k^{\text{applied}}$ are obtained by downsampling the 30\,Hz log at the solver boundaries. Angle residuals are wrapped to $[-\pi,\pi]$. $u^{\text{applied}}$ is the post-threshold control actually used by Robotarium, so the resulting residual $d_k$ captures true model mismatch, actuation, and sensing imperfections between the solver and the real system.

\subsection{Ellipsoidal set fitting for NRTO}
NRTO uses an ellipsoidal uncertainty set for the stacked disturbance vector
\begin{equation}
\zeta := [d_0^\top,\, d_1^\top,\, \ldots,\, d_T^\top]^\top \in \mathbb{R}^{(T+1)n_x}.
\end{equation}
In our implementation, the uncertainty is represented as
\begin{equation}
\zeta \in \{\Gamma z:\; z^\top S z \le \tau\},
\end{equation}
with $S\succ 0$ and scalar radius $\tau>0$.
From repeated open-loop rollouts, we estimate the start-offset statistics for $d_0$ and the per-step process statistics for $d_{k>0}$. We set $\Gamma=I$ and build $S$ as a block-diagonal matrix:
\begin{equation}
S = \mathrm{blkdiag}(S_0, S_d, \ldots, S_d),
\end{equation}
where $S_0\approx \Sigma_0^{-1}$ and $S_d\approx \Sigma_d^{-1}$ are the inverses of the empirical covariances of $d_0$ and $d_{k>0}$, respectively. Finally, we choose $\tau$ so that the ellipsoid covers the observed rollouts with a safety margin. For each rollout $r$, we compute the Mahalanobis energy
\begin{equation}
e_r := \zeta_r^\top S\, \zeta_r,
\end{equation}
and set $\tau$ to a conservative value such as $\tau = \max_r e_r$.

\subsection{Real-world deployment with NRTO feedback}
Solving the NRTO problem with the estimated $(\Gamma,S,\tau)$ yields a nominal control sequence $\bar u_k$ and a disturbance-feedback gain sequence $K_k$.
During execution, we apply the affine disturbance-feedback policy:
\begin{equation}
u_k = \mathrm{clamp}\big(\bar u_k + K_k\, d_k\big),
\end{equation}
where $d_k$ is estimated online from consecutive pose measurements using the same residual definition above, and $\mathrm{clamp}(\cdot)$ enforces Robotarium velocity limits and wheel-speed thresholding.

\end{document}